**Title:** Performance assessment of ADAS in a representative subset of critical traffic situations[*]

**Authors:** Luigi Di Lillo[1,3], Andrea Triscari[1], Xilin Zhou[1], Robert Dyro[2], Ruolin Li[2] and Marco Pavone[2]

[1] Swiss Reinsurance Company, Ltd
[2] Stanford University
[3] Autonomous Systems Lab, Stanford University (research affiliate)

## Abstract

As a variety of automated collision prevention systems gain presence within personal vehicles, rating and differentiating the automated safety performance of car models has become increasingly important for consumers, manufacturers, and insurers. In 2023, Swiss Re and partners initiated an eight-month long vehicle testing campaign conducted on a recognized UNECE type approval authority and Euro NCAP accredited proving ground in Germany. The campaign exposed twelve mass-produced vehicle models and one prototype vehicle fitted with collision prevention systems to a selection of safety-critical traffic scenarios representative of United States and European Union accident landscape. In this paper, we compare and evaluate the relative safety performance of these thirteen collision prevention systems (hardware and software stack) as demonstrated by this testing campaign. We first introduce a new scoring system which represents a test system's predicted impact on overall real-world collision frequency and reduction of collision impact energy, weighted based on the real-world relevance of the test scenario. Next, we introduce a novel metric that quantifies the *realism* of the protocol and confirm that our test protocol is a plausible representation of real-world driving. Finally, we find that the prototype system in its pre-release state outperforms the mass-produced (post-consumer-release) vehicles in the majority of the tested scenarios on the test track.

## Introduction

In recent years, developments in Advanced Driver Assistance Systems (ADAS) have led to broad shifts in the vehicle's role in road safety. While vehicle safety has historically focused on crashworthiness (i.e., passive safety) --a vehicle's physical resistance to damage upon impact--, ADAS instead intends to play a proactive role by altering the trajectory of the vehicle itself prior to a potential collision, achieved through functionality such as Automated Emergency Breaking (AEB). These advancements show promise in mitigating the potential severity of a collision and reducing the probability of collisions occurring in the first place [1].

AEB systems are engineered to recognize impending crashes and autonomously engage the brakes to avert or lessen the impact of an accident. Recognizing their significance, AEB systems have been mandated by numerous legislative authorities, with a compulsory integration into new models across the EU from May 2022 [2]. Likewise, in the US, the National Highway Traffic Safety Administration (NHTSA) has established a new Federal Motor Vehicle Safety Standard that will require AEB to be fitted standard in every new passenger vehicle and light truck by September 2029 [3].

A key dependency in the continued development, deployment, and acceptance of such technologies is the ability to reliably evaluate and compare the performance of ADAS systems. In recent years, national new car assessment programs (NCAPs) have been an important standardized reference for consumers. In particular, ADAS track testing protocols, which involve subjecting an ADAS-equipped vehicle to a set of collision scenarios on a closed track, have been successfully employed by the European organization of NCAP, Euro NCAP, since 2014 [4].

However, as such, current standardized scenario-testing protocols also face several challenges or limitations, namely in the selection and validation of safety-relevant scenarios and the interpretability of the scoring framework. To address such limitations, in 2023, Swiss Re and partners initiated an eight-month long testing campaign, conducted at a Euro NCAP accredited proving ground in Germany. The campaign exposed thirteen vehicle models fitted with collision prevention systems to a selection of safety-critical traffic scenarios relevant to both the United States and European Union accident statistics. The goal of this campaign was to overcome existing limitations facing current vehicle testing protocols, to introduce a comprehensive and safety-critical selection of scenarios, the validation of the scenario selection using a new realism metric, interpretable performance comparisons, and new measures.

---

[*] This study was funded by Luminar, which was not involved in the study design, data collection, analysis and interpretation of data, and writing of the report.

This paper reports on the results of the comparative assessment of the safety performance of the thirteen AEB systems. We first outline our paper's primary contributions, which include the definition of a new testing protocol, the definition and implementation of a new realism metric for validating the tested scenario, and a scoring framework which aims to represent relative safety performance between vehicles. Next, we introduce the testing campaign in further detail, as well as the scenario selection and validation and scoring methodologies. Finally, we report the relative safety performances of the thirteen AEB systems.

**Defining Safety Critical Scenarios**

In scenario-testing, testing protocols should reflect traffic situations which are *safety-critical*, which entails testing vehicles on the collision scenarios which most impact road safety today. For example, the current scenario-testing standard Euro NCAP defines its scenarios primarily by selecting the most *frequent* scenarios in the real world. However, considering frequency alone results in the exclusion of rarer but higher-severity collision scenarios, leaving a sizeable gap in evaluating the safety impact of ADAS. To account for an ADAS system's performance in higher-severity scenarios, we introduce a testing protocol by selecting scenarios based on a relevance score, derived from both the scenario's *frequency* (how often it occur) and *severity* (the average level of material damage).

**Quantifying Realism of Tested Scenarios**

A key consideration for the evaluation of ADAS systems is whether the scenarios used to stress test a software stack are *realistic*, that is, are plausible instantiations of driving scenarios that *could* happen in the real world. Generating realistic driving scenarios is a notoriously challenging task, as it entails reasoning about the nuanced behavior humans (e.g., drivers, bikers, and pedestrians) could display on the road. In the recent past, a number of works have addressed the problem of generating realistic driving scenarios, using techniques such as deep learning [5], and even leveraging large language models to provide grounding in police crash reports [6].

However, a standardized method for *quantifying* the realism of these generated scenarios, along with the corresponding responses provided by the system-under-test is, arguably, lacking. To address this gap, in a recent collaboration between Swiss Re and Stanford, we have developed a **realism score**, which makes use of behavior models trained on massive amounts of driving data to assess the realism of a generated scenario.

**Evaluating Relative Safety Outcomes Between Vehicles**

While safety ratings [7] and simple ADAS performance rankings [8] provide valuable preliminary insights and are readily available today to consumers, a key consideration for scoring frameworks is their ability to *quantify the relative safety outcomes* between vehicles tested. This is particularly relevant for parties like insurers who aim to quantify the potential insurance losses of one vehicle over another, as well as for safety impact researchers evaluating the impact of AEB technologies on road safety [9, 10].

However, quantifying relative safety outcomes via test track testing and other non-real-world testing (e.g. simulation) faces challenges. Scoring track test performance entails translating vehicle response data generated on the track into safety scores. To exceed the depth and quality of insights provided by current rankings and ratings, this translation must be designed to accurately represent real-world outcomes such as injury risk or economic losses so that the calculated relativities between scores are meaningful and interpretable. As such, currently available track testing scoring methodologies are limited in their ability to go beyond a ranking or rating system [11].

To address this gap, we capture performance relativities by introducing two measures of real-world safety outcomes: collision frequency reduction—the potential to reduce the number of collisions--and mitigation power—the expected reduction of energy transfer in a collision.

## Methodology

This paper introduces several new methodologies as part of a comprehensive mixed-methods comparative research design to evaluate the safety performance of thirteen AEB systems (Table 1). Specifically, the research involves track testing and the related development and implementation of the testing protocol, quantifying the realism of the tested scenarios, and finally the scoring framework to produce performance relativities across the thirteen tested vehicles.

**Track Testing**

The testing campaign was carried out on the Proving Grounds (PG) of an accredited NCAP organization (Euro NCAP | Euro NCAP – Members and Test Facilities), responsible for testing vehicles and developing new test protocols on behalf of NCAP. It is an engineering company, headquartered in Germany, which provides independent services to organizations active in the Automotive ecosystem. The tests object of this study was executed by this organization trained personnel and safety engineers, following full NCAP standardized procedures and approved soft targets.

*Vehicles Tested*

We tested a total of thirteen vehicles fitted with AEB functionality from the following makes: BMW, Chevrolet, Ford, Honda, Hyundai, Mercedes Benz, Nissan, Tesla, Toyota and Volkswagen. The vehicles have been anonymized and each assigned with an ID (Table 1). The ID consists of a number (i.e., 1, 2…11) and, for different models of the same make, an additional capital letter (i.e., A). Twelve of these vehicles were readily available to consumers at the time of testing, while one of the vehicles, ID 6, is a prototype fitted solely with LIDAR sensor and related system software stack technology. The vehicles were chosen based on their expected top safety performance and each was equipped with the full and latest set of safety features specifications.

AEB technologies, along with different ADAS, rely on numerous sensors and their associated software to avert crashes (Table 1). Each type of sensor has its own advantages and disadvantages related to range, field of view, spatial resolution, sensitivity to external factors, etc. [12]. Normally, a combination of different type of sensors allows the system to be more robust by combining their advantages and allowing redundancies to be implemented. Some manufactures choose to have only one type of sensor, relying on software development to overcome the sensor's limitations. This study covered a broad range of different front sensors suite currently being implemented by manufacturers.

| ID | Model Year | Front Sensors Suite | | | |
|---|---|---|---|---|---|
| | | Radar[1] | Corner Radars | Camera[2] | LIDAR[3] |
| 1A | 2022 | ✓ | ✓ | ✓ | |
| 1B | 2023 | ✓ | ✓ | ✓ | |
| 2 | 2023 | ✓ | | ✓ | |
| 3 | 2023 | ✓ | | ✓ | |
| 4 | 2023 | ✓ | ✓ | ✓ | |
| 5 | 2022 | ✓ | | ✓ | |
| 6 | - | | | | ✓ |
| 7A | 2023 | ✓ | ✓ | ✓ | |
| 7B | 2023 | ✓ | ✓ | ✓ | |
| 8 | 2022 | ✓ | | ✓ | |
| 9 | 2022 | | | ✓ | |
| 10 | 2023 | ✓ | | ✓ | |
| 11 | 2022 | ✓ | | ✓ | |

Table 1: The pool of tested vehicles consisted of twelve production vehicles from model years 2022–2023 and one prototype vehicle, ID 6.

---

[1] Radar systems are proficient in detecting objects over long distances and through harsh weather, yet they lack spatial resolution.
[2] Cameras, which capture detailed imagery, have limitations tied to visibility and blockages, akin to the limitations of the human eye.
[3] LIDARs create precise three-dimensional maps, filling the gap between radar's extensive detection capabilities and the camera's intricacy. On the other hand, they struggle under certain meteorological conditions such as rain and snow.

*Definition of the Testing Protocol*

To construct the testing matrix, we catalogued real-world collision scenarios using natural language processing on Swiss Re proprietary accident databases, and then selected scenarios by scoring them for *safety relevance*. The safety relevance of the scenarios, which are parametrized and clustered in a finite number of groupings composing the test protocol (see table 2 for an overview of this study's test protocol, and Appendix A for an in-depth view), is defined according to their real-world occurrence rate (i.e., frequency) and severity (i.e., injury potential and material damage potential[4]) level. Both the occurrence rate and the severity level represent their statistical significance. The test protocol utilized in this study is a subset of the Swiss Re's internally developed testing protocol, which is a scenario-based and location dependent test protocol.

Selected collision scenarios are organized via collision scenario groups, as seen in Table 2. In addition, each collision scenario can be tested at a range of scenario settings, which include the speed of the vehicle under test (VUT), the overlap between the VUT and the soft target (TG), and the simulated time of day. In sum, there are three scenario groups, thirty (if accounting for *opposite* and *same* direction turning scenarios) collision scenarios, and 224 possible tests overall. For the full form of the abbreviations, please refer to Appendix A.

To demonstrate how to read the below table, consider the Scenario Group Car to Car (C2C) and its scenario Car-to-Car Rear moving (CCRm). Our proposed test protocol for CCRm sets a range of speed for the VUT starting at 55 km/h (lower bound) to 125 km/h (upper bound), with the test speed of the VUT set at incremental steps of 10 km/h for a total of eight possible VUTs. The TG speed (in this case, a dummy NCAP-approved soft vehicle) is fixed at 20 km/h. The overlap refers to the degree of geometrical offset between the VUT, which starts 10% and moves to 50%, 75%, and 100% in daylight conditions for a total of four settings. In night conditions, the only overlap tested is 100%. In total, there are thirty-two test settings for CCRm in daylight conditions, and eight test settings for CCRm in night conditions.

In addition, table 2 offers a comparison between our test protocol and Euro NCAP AEB functionality test protocol. This comparison highlights that:
- The range of VUT's testing speeds of our test protocol is consistently shifted towards higher speeds than Euro NCAP's.
- The range of TG's testing speeds in our test protocol is shifted towards lower speeds than Euro NCAP's for two scenarios while it remains like Euro NCAP's for all the others.
- The *overlap* value is pretty much similar between the two protocols, except for two scenarios where our protocol appears to cover extra offsets.
- The *speed steps*, which represent the speed increment within a given scenario, appears to be wider, thus less refined, in our test protocol than Euro NCAP's.
- Finally, specifically for the AEB functionality, our test protocol covers overall a wider range of tests than Euro NCAP's.

---

[4] Insurance severity is a term encompassing injury potential, material damage potential and costs associated to both. In this work, any reference to severity is solely related to injury potential and material damage potential. Costs are not accounted for in any of our considerations or analyses.

| Scenario Group | Scenario[5] | Test Protocol | | | | | EuroNCAP (only AEB) | | | | |
|---|---|---|---|---|---|---|---|---|---|---|---|
| | | VUT speed [km/h] | TG Speed [km/h] | Speed steps [km/h] | Overlap [%] | Day Night | VUT speed [km/h] | TG Speed [km/h] | Speed steps [km/h] | Overlap [%] | Day Night |
| C2C | CCRs | 55-125 | - | 10 | 10/50/75/100 | Day | 10-80 | - | 5 | 50/75/100 | Day |
| | | 55-125 | - | 10 | 100 | Night | - | - | - | - | - |
| | CCRm | 55-125 | 20 | 10 | 10/50/75/100 | Day | 30-80 | 20 | 5 | 50/75/100 | Day |
| | | 55-125 | 20 | 10 | 100 | Night | - | - | - | - | - |
| | CCFtap | 15-25 | 35-55 | 10 | 50 | Day | 10-20 | 30/45/60 | 5 | 50 | Day |
| | | 15-25 | 35-55 | 10 | 50 | Night | - | - | - | - | - |
| | CCscp left | 35-95/35-65 | 15/35 | 10 | 25 | Day | 0/20-30 | 20-60 | 10 | 25 | Day |
| | | 35-95/35-65 | 15/35 | 10 | 25 | Night | - | - | - | - | - |
| | CCscp right | 35-95/35-65 | 15/35 | 10 | 25 | Day | - | - | - | - | - |
| | | 35-95/35-65 | 15/35 | 10 | 25 | Night | - | - | - | - | - |
| C2VRU | CPFA | - | - | - | - | - | 10-60 | 8 | 5 | 50 | Day |
| | | - | - | - | - | - | 10-60 | 8 | 5 | 50 | Night |
| | CPNA | - | - | - | - | - | 10-60 | 5 | 5 | 25/75 | Day |
| | | - | - | - | - | - | 10-60 | 5 | 5 | 25/75 | Night |
| | CPNAO | 45-75 | 5 | 10 | 50 | Day | | | | | |
| | | 45-75 | 5 | 10 | 50 | Night | | | | | |
| | CPLA | 45-85 | 5 | 10 | 50 | Day | 20-60 | 5 | 5 | 50 | Day |
| | | 45-85 | 5 | 10 | 50 | Night | - | - | - | - | - |
| | CPLAs | 45-85 | - | 10 | 50 | Night | - | - | - | - | - |
| | CPTA Farside | 15-35 | 5 | 10 | 50 | Day | 10-20 | 5 | 5 | 50 | Day |
| | | 15-35 | 5 | 10 | 50 | Night | - | - | - | - | - |
| | CPTA Nearside | 15-25 | 5 | 10 | 50 | Day | 10 | 5 | - | 50 | Day |
| | | 15-25 | 5 | 10 | 50 | Night | - | - | - | - | - |
| | CPNCO | 45-75 | 5 | 10 | 50 | Day | 10-60 | 5 | 5 | 50 | Day |
| | | 45-75 | 5 | 10 | 50 | Night | 10-60 | 5 | 5 | 50 | Night |
| | CPNCO Group | 45-75 | 5 | 10 | 50 | Day | - | - | - | - | - |
| | CPTC Farside | 15-35 | 5 | 10 | 50 | Day | - | - | - | - | - |
| | CPTC Nearside | 15-25 | 5 | 10 | 50 | Day | - | - | - | - | - |
| | CBFA | - | - | - | - | - | 10-60 | 20 | 5 | 50 | Day |
| | CBNA | 45-75 | 15 | 10 | 50 | Day | 10-60 | 15 | 5 | 50 | Day |
| | CBNAO | 45-85 | 10 | 10 | 50 | Day | 10-60 | 10 | 5 | 50 | Day |
| | CBLA | - | - | - | - | - | 25-60 | 15 | 5 | 50 | Day |
| | CBTA Farside | 15-35 | 15 | 10 | 50 | Day | 10-20 | 15 | 5 | 50 | Day |
| | CBTA Nearside | 15-25 | 15 | 10 | 50 | Day | 10 | 15 | - | 50 | Day |
| | Bobbycar | 45-75 | 5 | 10 | 50 | Day | - | - | - | - | - |
| | Scooter | 35-65 | 5/10/15 | 10 | 50 | Day | - | - | - | - | - |
| C2O | Pallet | 55-105 | - | 10 | 50 | Day | - | - | - | - | - |
| | | 45-105 | - | 10 | 50 | Night | - | - | - | - | - |
| | Tire | 55-105 | - | 10 | 50 | Day | - | - | - | - | - |
| | | 45-105 | - | 10 | 50 | Night | - | - | - | - | - |

Table 2: Detailed mapping of Euro NCAP (AEB) test protocol vs. our test protocol. Ranges are identified by the "-" character, while the "/" character is used to separate discrete test speeds, speed ranges, or overlaps.

---

[5] For simplification, we have omitted the direction of the pedestrian (same or opposite compared to the oncoming vehicle) in VRU turning scenarios.

*Test Protocol Procedure*

Each scenario is tested at an incrementally greater difficulty until the AEB fails to activate in response to the oncoming collision. Specifically, the tests were carried out starting with the lowest velocity specified for the scenario in the protocol. If the AEB activates, the test proceeds with the VUT speed increased by 10km/h. The testing continues until there is no response from the AEB or until the maximum speed of the protocol is reached. A similar approach was used for daylight vs. night scenarios. Namely, when an AEB displayed no response on a daylight test, the corresponding night test was not executed and judged as failed. For each test, a failure is defined as the complete lack of intervention of the automatized braking system. Table 2C in Appendix C shows the details of this procedure on the number of executed tests per each vehicle.

In addition, for any scenario settings that are not currently adopted by existing standardized testing protocols (e.g. Euro NCAP, see table 2 for a comparison), a "pre-test" was conducted prior to carrying out the incremental testing protocol defined above. Pre-tests are administered at a lower speed outside of the scenario's speed range defined by the protocol and is conducted to check for an AEB response without risking potential damage to the test equipment or the vehicle itself. In the case of a positive response from the AEB in the pre-test, the incremental testing procedure was then carried out. In the case of no response, the vehicle is judged as having failed the scenario.

**Quantifying Test Scenario Realism**

To quantify the "realism" of the tested scenarios, our approach leverages the implicit knowledge encoded within trained behavior prediction models. We assume that the optimal deep-learned driving behavior model parameters represent a close approximation of "data-aligned" realism fitted to the training data distribution. Deviations from these optimal parameters, therefore, correspond to decreasing levels of realism.

To develop the realism metrics, we make use of matrix sketching methods to construct a low-rank Laplace approximation of the posterior distribution of the parameters of any deep-trained behavior model (e.g., recurrent or transformer model), thereby allowing the mapping of directions in the learned parameter space to a likelihood score. By leveraging such a probabilistic interpretation, we can align a generated behavior with an observed (and thus real) behavior through optimization and find the maximum likelihood score that corresponds to the generated behavior. We refer to such a score as the "realism score." It is important to highlight that this approach is clearly just an approximation (please see limitation section for further details). However, this technique offers a valuable solution, replacing the subjective notion of realism with a measurable and scalable notion of *data-alignment*, measured within the parameter space of a trained behavior model. A more detailed description of the approach can be found in our pre-print [13].

**Scoring Framework**

To evaluate the safety impact and compare the safety performance of the tested pool of ADAS-equipped vehicles, we introduce two metrics: Frequency Score and Mitigation Power Score, defined below.

**Frequency score**: Being $\mathbb{P}[acc|passive]$ the overall probability of a vehicle without any ADAS fitment to be involved in a real-world accident, the Frequency Score (**FS**) measures the percentage reduction that a vehicle **x**, because of ADAS intervention, would cause to $\mathbb{P}[acc|passive]$.

In formula it reads:

$$FS = score_{freq}(x) \coloneqq 1 - \frac{\mathbb{P}[acc \mid veh = x]}{\mathbb{P}[acc \mid passive]} \quad (1)$$

Where $\mathbb{P}[acc|veh = x]$ represents the overall probability that vehicle **x** equipped with ADAS is involved in a real-world accident.

(1) can be rewritten as:

$$\mathbb{P}[acc \mid veh = x] = \left(1 - score_{freq}(x)\right) \mathbb{P}[acc \mid passive] \quad (2)$$

Equation (2) makes it evident that the **FS** measures the actual reduction of the overall probability of collision, with a higher score representing a lower probability of collision. By definition, a passive vehicle has a null frequency score.

Our scenario-based testing is organized in three Scenario Groups and a set of accident scenarios **s**, each assigned to the Groups (table X). To define a **FS** for each **s**, we make the following assumptions:

a) All the vehicles have the same probability to be involved in the configuration (i.e., a given set of speeds, offsets) of an accident scenario **s**.
b) If a vehicle is without ADAS, the accident will take place with probability one since no actual active features can intervene to avoid it.
c) A vehicle with ADAS, instead, would in principle be able to reduce the probability of incurring in an accident for a given **s** due to its advanced sensors and software stack system.
d) This reduction is measured by the proprietary Swiss Re model that yields the $\mathbb{P}[acc \mid veh = x, scen = s]$, which indicates the reduced probability of collision in scenario **s** due to the ADAS intervention of vehicle **x**.
e) The performance of **x** in each scenario are then aggregated according to the Swiss Re proprietary's statistical relevance of each **s** to obtain the overall reduced probability of accident $\mathbb{P}[acc \mid veh = x]$.

Equation 1 results from the aggregation of the reduction of probabilities the technology brings for each single tested scenario. As a result, we formally define the following scenario frequency score:

$$score_{freq}(x|s) := 1 - \frac{\mathbb{P}[acc \mid veh = x, scen = s]}{\mathbb{P}[acc \mid passive, scen = s]} \quad (3)$$

Where, $\mathbb{P}[acc \mid passive, scen = s]$ is the probability of a vehicle without ADAS to incur in an accident, conditioned to be in the configuration of a given **s**. Following assumption (a) we will set $\mathbb{P}[acc \mid passive, scen = s] = 1$ for any scenario **s**.

Analogously to (1), (3) can be rewritten as:

$$\mathbb{P}[acc \mid veh = x, scen = s] = \left(1 - score_{freq}(x|s)\right) \mathbb{P}[acc|passive, scen = s] \quad (4)$$

**Mitigation Power Score (MPS)**: This score measures the percentage reduction (mitigation) of the impact power $\mu_{pow}$ the vehicle **x** is expected to achieve in the scenario **s**, due to the AEB intervention. The score is computed with respect to a vehicle without any active safety feature (i.e., passive):

$$MPS = score_{MP}(x|s) = 1 - \frac{\mathbb{E}[\mu_{pow}|veh = x, scen = s]}{\mathbb{E}[\mu_{pow}|passive, scen = s]} \quad (5)$$

In equation (5), $\mu_{pow}$ considers the speed of the impact, the masses of the vehicles involved, as well as the geometry of the collision[6]. The expectation value is computed over the probability distribution of the static and dynamical settings which influence $\mu_{pow}$ and characterize the scenario **s**.

Analogously to (4), the score reads:

$$\mathbb{E}[\mu_{pow}|veh = x, scen = s] = (1 - score_{MP}(x|s))\mathbb{E}[\mu_{pow}|passive, scen = s] \quad (6)$$

Equation (6) represents the percentage reduction to the expected impact power of vehicle **x** in the scenario **s**, with a given configuration, that the technology can bring compared to a passive vehicle.

---

[6] The explicit formulations of the **P** and $\mu_{pow}$ is Swiss Re's intellectual property.

*Scores Aggregation*

We implement an aggregated scoring approach, which matches the hierarchical organization of the scenario-based testing protocol introduced in table 2. The performance of each vehicle is first defined at *scenario level* to provide an estimation of the system's performance within a specific collision scenario across the given settings. *Scenario level* scores are then aggregated, according to the statistical relevance of each scenario in the real-word accident statistics, to produce *scenario group* (SG) scores. Scenario group scores provide an overview of the expected improvement in the safety of the vehicle within a broader category of collisions.

The aggregation is described as follows:

Assuming that the C2VRU *scenario group* is composed of two scenarios, namely CPLA day and CPNCO day, we would compute the **FS** and **MPS** scores of vehicle **x** first at the *scenario level s*:

$$score_{freq}(x|s = CPNCO_{day}), \quad score_{freq}(x|s = CPLA_{day})$$

$$score_{MP}(x|s = CPNCO_{day}), \quad score_{MP}(x|s = CPLA_{day})$$

Subsequently, the SG scores are computed as follows:

$$score_{freq}^{C2VRU}(x) = \sum_{s \in S(C2VRU)} W[s] \cdot score_{freq}(x|s) \quad (7)$$

$$score_{MP}^{C2VRU}(x) = \frac{\sum_{s \in S(C2VRU)} W[s] \cdot \mu_{pow}(s) \cdot score_{MP}(x|s)}{\sum_{s \in S(C2VRU)} W[s] \cdot \mu_{pow}(s)} \quad (8)$$

Where **W[s]** takes into account the statistical relevance of scenario **s** in the population of scenarios in the selected group, and **S** is the scenarios' associating function that associates to each **SG** the set of scenarios belonging to the group.

With reference to the above example, we have:

$$S(C2VRU) = \{CPLA_{day}, CPNCO_{day}\}$$

From equations (7) and (8) we obtain:
1. The FS, at group level, indicating the overall expected reduction of probability of accidents for the whole set of scenarios in that specific group for vehicle **x**.
2. The MPS, at group level, indicates the extent to which vehicle **x** is expected to mitigate the overall impact power for the whole set of scenarios in that specific group.

*Relative Scores*

We will now define a metrics to relate the FS and MPS values of each single vehicle to any other vehicle present in the test pool.

The goal is to understand the level of relative improvement/worsening that a certain safety technology (i.e., combination of hardware and software stack) is able to bring with respect to a different implementation of the same safety feature.

To evaluate the relative performance between vehicles, we additionally formally define the relative performance $rel_M(x|y)$ of vehicle **x** with respect to vehicle **y** according to the metrics **M**:

$$rel_M(x|y) = \frac{score_M(x)}{score_M(y)} - 1 \quad (9)$$

$$for\ M : Metrics\ \in \{freq, MP\}$$

Similarly to (1) and (4), (9) can be formally rewritten as:

$$score_M(x) = [1 + rel_M(x|y)]\ score_M(y) \quad (10)$$

Equation (10) underpins the meaning of the relative score, namely:

1. **Relative Frequency Score:** It indicates the percentage decrease or increase in the collision probability of vehicle **x** with respect to vehicle **y.** For example, if $rel_{freq}(x|y)$ = -20%, then vehicle **x** is expected to prevent 20% fewer collisions than vehicle **y** for a given scenario or scenario group.
2. **Relative Mitigation Power Score:** It represents how much impact power vehicle **x** is expected to mitigate with respect to vehicle **y**. For example, if $rel_{MP}(x|y)$ = -20%, vehicle **x** is expected to mitigate 20% less impact power than vehicle **y** for a given scenario or scenario group.

## Results and Discussions

*Scenario Realism*

We first measured the realism score for the scenarios and vehicle responses considered as part of this study and found that all considered scenarios are highly realistic in the sense of our non-dimensional realism score (mean = 0.002354, standard deviation = 0.001221, where values above ~0.01 indicate largely unrealistic scenarios as determined experimentally in [13]). A qualitative human validation step on a small sample of the scenarios provided support to the validity of this assessment.

*Comparative Assessment*

The following eight tables (Tables 3 - 10) show the comparative performance of the tested vehicles across two scenario groups ("car to car," "car to vulnerable road user"), two metrics (frequency score and mitigation power score), and two scenario aggregation (i.e. statistical weighting) systems (United States and European Union) which are each representative of the respective region's driving accident statistics[7].

The percentage in each cell indicates the *relative* difference between the scores of vehicle ID x (row) and vehicle ID y (column), normalized in terms of the score of vehicle y[8]. The vehicles IDs are ranked in terms of performance, from the best performer to the worst performer. The colored grading system similarly visualizes this relativity, with red used for a relatively weaker performance, and green for a relatively better performance. Given that the performance scores are presented as relativities, the values in the cells do not represent an absolute performance of the tested pool of vehicles in preventing or mitigating accidents. Namely, a substantial % difference in x-y vehicles' combinations performance, does not mean that x or y is very unsafe but that one of them in comparison to the other is estimated to be much safer.

To further demonstrate the interpretation of the following tables, consider Table 3, which shows a comparison of the frequency scores for the Car-to-Car SG, with individual scenarios weighted according to EU accident statistics. According to the two cells comparing the ID 7B and ID 4 performances, the former is expected to prevent ~39% more collisions than the latter in that SG in the EU. Conversely, the ID 4 is expected to experience ~28% more collisions in the same SG than the ID 7B.

Similarly, consider Table 5, which portrays a comparison of the mitigation power scores for the Car-to-Car scenario group, with individual scenarios weighted according to EU accident statistics. According to the two cells comparing ID 7B and ID 4 performances, ID 7B is expected to mitigate ~20% more energy transfer during a collision than the ID 4 in in the EU. Conversely, the ID 4 is expected to transfer ~17% more impact energy during a collision than the ID 7B.

A cell which contains the value *inf* indicates that the vehicle y (column) is not expected to avoid collisions in the scenario group.

---

[7] At this level of granularity, we aggregate across test settings (e.g., light conditions, speed, and offsets) to show performance at the scenario group level. To measure performance specific to test settings, such as a vehicle's low light performance in comparison to another vehicle's, a different aggregation approach would be required.
[8] In presenting relative performance, the matrix demonstrates a null-diagonal behavior and is also asymmetric due to the x-y asymmetry of the relative score definition.

| Relative Frequency Scores, Car-to-Car (European Union) | | | | | | | | | | | | | |
|---|---|---|---|---|---|---|---|---|---|---|---|---|---|
| | ID 6 | ID 7B | ID 4 | ID 9 | ID 1A | ID 11 | ID 8 | ID 7A | ID 10 | ID 1B | ID 2 | ID 5 | ID 3 |
| ID 6 | 0.00% | 12.86% | 56.78% | 74.27% | 81.55% | 92.44% | 89.52% | 129.58% | 169.26% | 189.89% | 1799.63% | 1799.63% | inf |
| ID 7B | -11.39% | 0.00% | 38.92% | 54.41% | 60.87% | 70.52% | 67.92% | 103.43% | 138.59% | 156.87% | 1583.22% | 1583.22% | inf |
| ID 4 | -36.22% | -28.02% | 0.00% | 11.15% | 15.80% | 22.75% | 20.88% | 46.44% | 71.75% | 84.90% | 1111.65% | 1111.65% | inf |
| ID 9 | -42.62% | -35.24% | -10.03% | 0.00% | 4.18% | 10.43% | 8.75% | 31.74% | 54.51% | 66.35% | 990.08% | 990.08% | inf |
| ID 1A | -44.92% | -37.84% | -13.64% | -4.01% | 0.00% | 6.00% | 4.39% | 26.46% | 48.31% | 59.68% | 946.35% | 946.35% | inf |
| ID 11 | -48.04% | -41.35% | -18.53% | -9.44% | -5.66% | 0.00% | -1.52% | 19.30% | 39.92% | 50.64% | 887.13% | 887.13% | inf |
| ID 8 | -47.23% | -40.45% | -17.27% | -8.05% | -4.20% | 1.54% | 0.00% | 21.14% | 42.08% | 52.97% | 902.36% | 902.36% | inf |
| ID 7A | -56.44% | -50.84% | -31.71% | -24.09% | -20.92% | -16.18% | -17.45% | 0.00% | 17.28% | 26.27% | 727.43% | 727.43% | inf |
| ID 10 | -62.86% | -58.09% | -41.77% | -35.28% | -32.58% | -28.53% | -29.62% | -14.74% | 0.00% | 7.66% | 605.49% | 605.49% | inf |
| ID 1B | -65.50% | -61.07% | -45.92% | -39.89% | -37.37% | -33.62% | -34.63% | -20.80% | -7.12% | 0.00% | 555.28% | 555.28% | inf |
| ID 2 | -94.74% | -94.06% | -91.75% | -90.83% | -90.44% | -89.87% | -90.02% | -87.91% | -85.83% | -84.74% | 0.00% | 0.00% | inf |
| ID 5 | -94.74% | -94.06% | -91.75% | -90.83% | -90.44% | -89.87% | -90.02% | -87.91% | -85.83% | -84.74% | 0.00% | 0.00% | inf |
| ID 3 | -100.00% | -100.00% | -100.00% | -100.00% | -100.00% | -100.00% | -100.00% | -100.00% | -100.00% | -100.00% | -100.00% | -100.00% | 0.00% |

*Table 3: Comparative matrix of relative frequency scores (FS) at the scenario group level "car to car", for the EU protocol. Each cell shows the relative FS between vehicle x (row) and vehicle y (column).*

| Relative Frequency Scores, Car-to-Car (United States) | | | | | | | | | | | | | |
|---|---|---|---|---|---|---|---|---|---|---|---|---|---|
| | ID 6 | ID 7B | ID 4 | ID 9 | ID 1A | ID 11 | ID 8 | ID 7A | ID 1B | ID 10 | ID 5 | ID 2 | ID 3 |
| ID 6 | 0.00% | 15.69% | 31.70% | 43.00% | 48.82% | 53.82% | 53.10% | 63.02% | 82.21% | 84.87% | 1329.99% | 1330.13% | inf |
| ID 7B | -13.56% | 0.00% | 13.84% | 23.61% | 28.64% | 32.96% | 32.34% | 40.91% | 57.50% | 59.80% | 1136.07% | 1136.19% | inf |
| ID 4 | -24.07% | -12.16% | 0.00% | 8.58% | 13.00% | 16.79% | 16.24% | 23.78% | 38.34% | 40.37% | 985.76% | 985.87% | inf |
| ID 9 | -30.07% | -19.10% | -7.90% | 0.00% | 4.07% | 7.56% | 7.06% | 14.00% | 27.41% | 29.28% | 899.97% | 900.07% | inf |
| ID 1A | -32.81% | -22.26% | -11.50% | -3.91% | 0.00% | 3.36% | 2.87% | 9.54% | 22.43% | 24.22% | 860.88% | 860.97% | inf |
| ID 11 | -34.99% | -24.79% | -14.38% | -7.03% | -3.25% | 0.00% | -0.47% | 5.98% | 18.46% | 20.19% | 829.68% | 829.77% | inf |
| ID 8 | -34.68% | -24.43% | -13.97% | -6.59% | -2.79% | 0.47% | 0.00% | 6.48% | 19.01% | 20.75% | 834.04% | 834.13% | inf |
| ID 7A | -38.66% | -29.03% | -19.21% | -12.28% | -8.71% | -5.65% | -6.09% | 0.00% | 11.77% | 13.40% | 777.18% | 777.27% | inf |
| ID 1B | -45.12% | -36.51% | -27.72% | -21.52% | -18.32% | -15.58% | -15.98% | -10.53% | 0.00% | 1.46% | 684.82% | 684.90% | inf |
| ID 10 | -45.91% | -37.42% | -28.76% | -22.65% | -19.50% | -16.80% | -17.18% | -11.82% | -1.44% | 0.00% | 673.52% | 673.60% | inf |
| ID 5 | -93.01% | -91.91% | -90.79% | -90.00% | -89.59% | -89.24% | -89.29% | -88.60% | -87.26% | -87.07% | 0.00% | 0.01% | inf |
| ID 2 | -93.01% | -91.91% | -90.79% | -90.00% | -89.59% | -89.24% | -89.29% | -88.60% | -87.26% | -87.07% | -0.01% | 0.00% | inf |
| ID 3 | -100.00% | -100.00% | -100.00% | -100.00% | -100.00% | -100.00% | -100.00% | -100.00% | -100.00% | -100.00% | -100.00% | -100.00% | 0.00% |

*Table 4: Comparative matrix of frequency relative frequency scores (FS) at the scenario group level "car to car", for the US protocol. Each cell shows the relative FS between vehicle x (row) and vehicle y (column).*

In the EU test protocol, the ID 6's ability to prevent collisions in car-to-car pre-collision scenarios on a test track outperforms that of all vehicles in the test pool by at least 10% (Table 3). A similar result is demonstrated in the US protocol; however, ID 6's performance gap is smaller across all comparisons except for that with ID 7B (Table 4). While this is an indicator of good performance by the prototype, it must be considered alongside additional aspects described in the below *Limitations and Considerations* section, namely that strong test track performance—although indicative of safety performance—does not necessarily translate to a comfortable driving experience once deployed on real roads.

In addition, between EU and US protocols, some interesting changes in performance across some vehicles' combinations might be observed. For example, when considering the ID 10 – ID 1B combination, the latter performs comparatively better on US roads than EU roads.

| Relative Mitigation Power Scores, Car-to-Car (European Union) | | | | | | | | | | | | | |
|---|---|---|---|---|---|---|---|---|---|---|---|---|---|
|  | ID 6 | ID 7B | ID 4 | ID 9 | ID 1A | ID 11 | ID 8 | ID 7A | ID 10 | ID 1B | ID 2 | ID 5 | ID 3 |
| ID 6 | 0.00% | 32.94% | 59.51% | 60.31% | 114.76% | 80.51% | 127.38% | 134.40% | 217.81% | 213.81% | 4651.12% | 373.98% | inf |
| ID 7B | -24.78% | 0.00% | 19.99% | 20.59% | 61.55% | 35.78% | 71.03% | 76.31% | 139.06% | 136.05% | 3473.83% | 256.53% | inf |
| ID 4 | -37.31% | -16.66% | 0.00% | 0.50% | 34.64% | 13.16% | 42.54% | 46.95% | 99.24% | 96.73% | 2878.52% | 197.14% | inf |
| ID 9 | -37.62% | -17.07% | -0.50% | 0.00% | 33.96% | 12.60% | 41.83% | 46.21% | 98.24% | 95.75% | 2863.63% | 195.66% | inf |
| ID 1A | -53.44% | -38.10% | -25.73% | -25.35% | 0.00% | -15.95% | 5.87% | 9.14% | 47.98% | 46.12% | 2112.27% | 120.70% | inf |
| ID 11 | -44.60% | -26.35% | -11.63% | -11.19% | 18.98% | 0.00% | 25.96% | 29.85% | 76.07% | 73.85% | 2532.07% | 162.58% | inf |
| ID 8 | -56.02% | -41.53% | -29.85% | -29.49% | -5.55% | -20.61% | 0.00% | 3.09% | 39.77% | 38.01% | 1989.55% | 108.46% | inf |
| ID 7A | -57.34% | -43.28% | -31.95% | -31.61% | -8.38% | -22.99% | -3.00% | 0.00% | 35.59% | 33.88% | 1926.96% | 102.21% | inf |
| ID 10 | -68.54% | -58.17% | -49.81% | -49.56% | -32.43% | -43.20% | -28.46% | -26.25% | 0.00% | -1.26% | 1394.94% | 49.14% | inf |
| ID 1B | -68.13% | -57.64% | -49.17% | -48.91% | -31.56% | -42.48% | -27.54% | -25.31% | 1.28% | 0.00% | 1414.02% | 51.04% | inf |
| ID 2 | -97.90% | -97.20% | -96.64% | -96.63% | -95.48% | -96.20% | -95.21% | -95.07% | -93.31% | -93.40% | 0.00% | -90.02% | inf |
| ID 5 | -78.90% | -71.95% | -66.35% | -66.18% | -54.69% | -61.92% | -52.03% | -50.55% | -32.95% | -33.79% | 902.39% | 0.00% | inf |
| ID 3 | -100.00% | -100.00% | -100.00% | -100.00% | -100.00% | -100.00% | -100.00% | -100.00% | -100.00% | -100.00% | -100.00% | -100.00% | 0.00% |

*Table 5: Comparative matrix of relative mitigation power scores at the scenario group level "car to car", for the EU protocol. Each cell shows the relative MPS between vehicle x (row) and vehicle y (column).*

| Relative Mitigation Power Scores, Car-to-Car (United States) | | | | | | | | | | | | | |
|---|---|---|---|---|---|---|---|---|---|---|---|---|---|
|  | ID 6 | ID 7B | ID 4 | ID 9 | ID 1A | ID 11 | ID 8 | ID 7A | ID 1B | ID 10 | ID 5 | ID 2 | ID 3 |
| ID 6 | 0.00% | 32.16% | 36.54% | 46.80% | 82.55% | 55.28% | 77.17% | 89.51% | 109.95% | 116.92% | 162.43% | 3387.57% | inf |
| ID 7B | -24.33% | 0.00% | 3.31% | 11.07% | 38.13% | 17.50% | 34.05% | 43.39% | 58.86% | 64.13% | 98.57% | 2538.88% | inf |
| ID 4 | -26.76% | -3.20% | 0.00% | 7.51% | 33.70% | 13.73% | 29.76% | 38.80% | 53.77% | 58.87% | 92.21% | 2454.31% | inf |
| ID 9 | -31.88% | -9.97% | -6.99% | 0.00% | 24.36% | 5.78% | 20.69% | 29.10% | 43.02% | 47.77% | 78.77% | 2275.80% | inf |
| ID 1A | -45.22% | -27.60% | -25.21% | -19.59% | 0.00% | -14.94% | -2.95% | 3.81% | 15.01% | 18.83% | 43.76% | 1810.46% | inf |
| ID 11 | -35.60% | -14.89% | -12.07% | -5.47% | 17.56% | 0.00% | 14.09% | 22.04% | 35.20% | 39.69% | 69.00% | 2145.93% | inf |
| ID 8 | -43.56% | -25.40% | -22.93% | -17.14% | 3.04% | -12.35% | 0.00% | 6.97% | 18.50% | 22.44% | 48.13% | 1868.52% | inf |
| ID 7A | -47.23% | -30.26% | -27.95% | -22.54% | -3.67% | -18.06% | -6.51% | 0.00% | 10.79% | 14.46% | 38.48% | 1740.32% | inf |
| ID 1B | -52.37% | -37.05% | -34.97% | -30.08% | -13.05% | -26.04% | -15.61% | -9.74% | 0.00% | 3.32% | 25.00% | 1561.14% | inf |
| ID 10 | -53.90% | -39.07% | -37.06% | -32.33% | -15.84% | -28.41% | -18.33% | -12.64% | -3.21% | 0.00% | 20.98% | 1507.77% | inf |
| ID 5 | -61.89% | -49.64% | -47.97% | -44.06% | -30.44% | -40.83% | -32.49% | -27.79% | -20.00% | -17.34% | 0.00% | 1228.94% | inf |
| ID 2 | -97.13% | -96.21% | -96.09% | -95.79% | -94.77% | -95.55% | -94.92% | -94.57% | -93.98% | -93.78% | -92.48% | 0.00% | inf |
| ID 3 | -100.00% | -100.00% | -100.00% | -100.00% | -100.00% | -100.00% | -100.00% | -100.00% | -100.00% | -100.00% | -100.00% | -100.00% | 0.00% |

*Table 6: Comparative matrix of relative mitigation power scores at the scenario group level "car to car", for the US protocol. Each cell shows the relative MPS between vehicle x (row) and vehicle y (column).*

For the mitigation of energy transfer in the event of an impact, in the EU protocol and car-to-car scenarios, the ID 6 prototype outperforms ID 7B and ID 4 by over 30% and almost 60%, respectively (Table 5). ID 7B and ID 4 are second and third best performers, outperforming any other vehicle in the test pool by ~20% and ~1%, respectively. Similarly to the frequency score relativities in Tables 3 and 4, ID 6's performance gap is smaller across all comparisons in the US protocol (Table 6) than EU's.

| Relative Frequency Scores, Car-to-Vulnerable Road User (European Union) | | | | | | | | | | | | | |
|---|---|---|---|---|---|---|---|---|---|---|---|---|---|
| | ID 6 | ID 9 | ID 1A | ID 10 | ID 7B | ID 7A | ID 8 | ID 4 | ID 1B | ID 3 | ID 11 | ID 2 | ID 5 |
| ID 6 | 0.00% | 42.51% | 55.14% | 62.61% | 94.55% | 118.12% | 336.74% | 341.24% | 733.90% | 824.34% | 920.01% | Inf | inf |
| ID 9 | -29.83% | 0.00% | 8.86% | 14.11% | 36.52% | 53.06% | 206.47% | 209.63% | 485.16% | 548.63% | 615.76% | Inf | inf |
| ID 1A | -35.54% | -8.14% | 0.00% | 4.82% | 25.41% | 40.60% | 181.52% | 184.42% | 437.52% | 495.82% | 557.48% | Inf | inf |
| ID 10 | -38.50% | -12.36% | -4.59% | 0.00% | 19.64% | 34.14% | 168.58% | 171.35% | 412.82% | 468.44% | 527.27% | Inf | inf |
| ID 7B | -48.60% | -26.75% | -20.26% | -16.42% | 0.00% | 12.11% | 124.48% | 126.79% | 328.62% | 375.10% | 424.28% | Inf | inf |
| ID 7A | -54.15% | -34.67% | -28.87% | -25.45% | -10.80% | 0.00% | 100.23% | 102.29% | 282.31% | 323.78% | 367.64% | Inf | inf |
| ID 8 | -77.10% | -67.37% | -64.48% | -62.77% | -55.45% | -50.06% | 0.00% | 1.03% | 90.93% | 111.64% | 133.55% | Inf | inf |
| ID 4 | -77.34% | -67.70% | -64.84% | -63.15% | -55.91% | -50.57% | -1.02% | 0.00% | 88.99% | 109.49% | 131.17% | Inf | inf |
| ID 1B | -88.01% | -82.91% | -81.40% | -80.50% | -76.67% | -73.84% | -47.63% | -47.09% | 0.00% | 10.85% | 22.32% | Inf | inf |
| ID 3 | -89.18% | -84.58% | -83.22% | -82.41% | -78.95% | -76.40% | -52.75% | -52.26% | -9.78% | 0.00% | 10.35% | Inf | inf |
| ID 11 | -90.20% | -86.03% | -84.79% | -84.06% | -80.93% | -78.62% | -57.18% | -56.74% | -18.25% | -9.38% | 0.00% | Inf | inf |
| ID 2 | -100.00% | -100.00% | -100.00% | -100.00% | -100.00% | -100.00% | -100.00% | -100.00% | -100.00% | -100.00% | -100.00% | 0.00% | 0.00% |
| ID 5 | -100.00% | -100.00% | -100.00% | -100.00% | -100.00% | -100.00% | -100.00% | -100.00% | -100.00% | -100.00% | -100.00% | 0.00% | 0.00% |

*Table 7: Comparative matrix of relative frequency scores at the scenario group level "car to vulnerable road user", for the EU protocol. Each cell shows the relative FS between vehicle x (row) and vehicle y column).*

| Relative Frequency Scores, Car-to-Vulnerable Road User (United States) | | | | | | | | | | | | | |
|---|---|---|---|---|---|---|---|---|---|---|---|---|---|
| | ID 6 | ID 4 | ID 1A | ID 7A | ID 10 | ID 8 | ID 3 | ID 9 | ID 1B | ID 11 | ID 7B | ID 2 | ID 5 |
| ID 6 | 0.00% | 25.28% | 28.22% | 57.44% | 68.09% | 69.49% | 72.06% | 258.71% | 305.10% | 373.22% | 774.96% | Inf | inf |
| ID 4 | -20.18% | 0.00% | 2.35% | 25.67% | 34.18% | 35.29% | 37.34% | 186.33% | 223.36% | 277.73% | 598.41% | Inf | inf |
| ID 1A | -22.01% | -2.30% | 0.00% | 22.79% | 31.09% | 32.18% | 34.19% | 179.75% | 215.93% | 269.06% | 582.37% | Inf | inf |
| ID 7A | -36.48% | -20.43% | -18.56% | 0.00% | 6.77% | 7.65% | 9.28% | 127.84% | 157.30% | 200.56% | 455.73% | Inf | inf |
| ID 10 | -40.51% | -25.47% | -23.72% | -6.34% | 0.00% | 0.83% | 2.36% | 113.40% | 141.00% | 181.52% | 420.52% | Inf | inf |
| ID 8 | -41.00% | -26.09% | -24.35% | -7.11% | -0.82% | 0.00% | 1.51% | 111.64% | 139.01% | 179.20% | 416.23% | Inf | inf |
| ID 3 | -41.88% | -27.19% | -25.48% | -8.49% | -2.30% | -1.49% | 0.00% | 108.48% | 135.45% | 175.03% | 408.53% | Inf | inf |
| ID 9 | -72.12% | -65.08% | -64.25% | -56.11% | -53.14% | -52.75% | -52.03% | 0.00% | 12.93% | 31.92% | 143.92% | Inf | inf |
| ID 1B | -75.31% | -69.07% | -68.35% | -61.13% | -58.51% | -58.16% | -57.53% | -11.45% | 0.00% | 16.81% | 115.98% | Inf | inf |
| ID 11 | -78.87% | -73.53% | -72.90% | -66.73% | -64.48% | -64.18% | -63.64% | -24.20% | -14.39% | 0.00% | 84.90% | Inf | inf |
| ID 7B | -88.57% | -85.68% | -85.35% | -82.01% | -80.79% | -80.63% | -80.34% | -59.00% | -53.70% | -45.92% | 0.00% | Inf | inf |
| ID 2 | -100.00% | -100.00% | -100.00% | -100.00% | -100.00% | -100.00% | -100.00% | -100.00% | -100.00% | -100.00% | -100.00% | 0.00% | 0.00% |
| ID 5 | -100.00% | -100.00% | -100.00% | -100.00% | -100.00% | -100.00% | -100.00% | -100.00% | -100.00% | -100.00% | -100.00% | 0.00% | 0.00% |

*Table 8: Comparative matrix of relative frequency scores at the scenario group level "car to vulnerable road user", for the US protocol. Each cell shows the relative FS between vehicle x (row) and vehicle y (column).*

The prototype's ability to prevent collisions in car-to-vulnerable road user (C2VRU) pre-collision scenarios on a test track outperforms that of all vehicles in the test pool by at least 40% given the EU protocol (Table 7), and at least 25% for the US protocol (Table 8).

In addition, there are several differences in the performance rankings between the US and EU protocols. For example, ID 9 is estimated showing ~72% more C2VRU collisions than ID 6 in the US, whereas that percentage drops to ~30% more C2VRU collisions than ID 6 in the EU. The differences in vehicles' performance between protocols also indicate that the frequencies of the individual C2VRU scenarios differ markedly between the two regions. This aspect is indicative of the different traffic dynamics in the different countries.

| Relative Mitigation Power Scores, Car-to-Vulnerable Road User (European Union) | | | | | | | | | | | | | |
|---|---|---|---|---|---|---|---|---|---|---|---|---|---|
|  | ID 6 | ID 9 | ID 1A | ID 10 | ID 7B | ID 7A | ID 8 | ID 4 | ID 1B | ID 3 | ID 11 | ID 2 | ID 5 |
| ID 6 | 0.00% | 48.44% | 43.40% | 49.99% | 72.17% | 66.63% | 109.71% | 36.13% | 88.28% | 304.33% | 255.51% | inf | 675.44% |
| ID 9 | -32.63% | 0.00% | -3.40% | 1.04% | 15.99% | 12.26% | 41.28% | -8.29% | 26.84% | 172.39% | 139.50% | inf | 422.40% |
| ID 1A | -30.26% | 3.52% | 0.00% | 4.60% | 20.07% | 16.20% | 46.24% | -5.07% | 31.30% | 181.97% | 147.92% | -inf | 440.77% |
| ID 10 | -33.33% | -1.03% | -4.40% | 0.00% | 14.79% | 11.10% | 39.82% | -9.24% | 25.53% | 169.57% | 137.02% | inf | 417.00% |
| ID 7B | -41.92% | -13.78% | -16.71% | -12.88% | 0.00% | -3.22% | 21.80% | -20.93% | 9.36% | 134.84% | 106.49% | inf | 350.39% |
| ID 7A | -39.99% | -10.92% | -13.94% | -9.99% | 3.32% | 0.00% | 25.85% | -18.30% | 12.99% | 142.65% | 113.35% | inf | 365.36% |
| ID 8 | -52.31% | -29.22% | -31.62% | -28.48% | -17.90% | -20.54% | 0.00% | -35.08% | -10.22% | 92.81% | 69.53% | inf | 269.77% |
| ID 4 | -26.54% | 9.04% | 5.34% | 10.18% | 26.47% | 22.40% | 54.05% | 0.00% | 38.31% | 197.01% | 161.15% | inf | 469.62% |
| ID 1B | -46.89% | -21.16% | -23.84% | -20.34% | -8.56% | -11.50% | 11.38% | -27.70% | 0.00% | 114.75% | 88.82% | inf | 311.85% |
| ID 3 | -75.27% | -63.29% | -64.54% | -62.90% | -57.42% | -58.79% | -48.13% | -66.33% | -53.43% | 0.00% | -12.07% | inf | 91.78% |
| ID 11 | -71.87% | -58.25% | -59.66% | -57.81% | -51.57% | -53.13% | -41.01% | -61.71% | -47.04% | 13.73% | 0.00% | inf | 118.12% |
| ID 2 | -100.00% | -100.00% | -100.00% | -100.00% | -100.00% | -100.00% | -100.00% | -100.00% | -100.00% | -100.00% | -100.00% | 0.00% | -100.00% |
| ID 5 | -87.10% | -80.86% | -81.51% | -80.66% | -77.80% | -78.51% | -72.96% | -82.44% | -75.72% | -47.86% | -54.15% | inf | 0.00% |

*Table 9: Comparative matrix of relative mitigation power scores at the scenario group level "car to vulnerable road user", for the EU protocol. Each cell shows the relative MPS between vehicle x (row) and vehicle y (column).*

| Relative Mitigation Power Scores, Car-to-Vulnerable Road User (United States) | | | | | | | | | | | | | |
|---|---|---|---|---|---|---|---|---|---|---|---|---|---|
|  | ID 6 | ID 4 | ID 1A | ID 7A | ID 10 | ID 8 | ID 3 | ID 9 | ID 1B | ID 11 | ID 7B | ID 2 | ID 5 |
| ID 6 | 0.00% | 22.63% | 41.74% | 64.89% | 81.18% | 108.74% | 84.94% | 326.71% | 71.63% | 101.07% | 314.87% | -inf | 3607.03% |
| ID 4 | -18.46% | 0.00% | 15.58% | 34.46% | 47.74% | 70.22% | 50.81% | 247.96% | 39.96% | 63.96% | 238.30% | -inf | 2922.87% |
| ID 1A | -29.45% | -13.48% | 0.00% | 16.34% | 27.83% | 47.27% | 30.48% | 201.06% | 21.09% | 41.86% | 192.70% | -inf | 2515.44% |
| ID 7A | -39.35% | -25.63% | -14.04% | 0.00% | 9.88% | 26.59% | 12.16% | 158.78% | 4.09% | 21.94% | 151.60% | -inf | 2148.16% |
| ID 10 | -44.81% | -32.31% | -21.77% | -8.99% | 0.00% | 15.21% | 2.08% | 135.52% | -5.27% | 10.98% | 128.98% | -inf | 1946.06% |
| ID 8 | -52.09% | -41.25% | -32.10% | -21.01% | -13.20% | 0.00% | -11.40% | 104.42% | -17.78% | -3.67% | 98.75% | -inf | 1675.91% |
| ID 3 | -45.93% | -33.69% | -23.36% | -10.84% | -2.03% | 12.87% | 0.00% | 130.73% | -7.19% | 8.72% | 124.33% | -inf | 1904.44% |
| ID 9 | -76.56% | -71.26% | -66.78% | -61.36% | -57.54% | -51.08% | -56.66% | 0.00% | -59.78% | -52.88% | -2.78% | -inf | 768.75% |
| ID 1B | -41.74% | -28.55% | -17.42% | -3.93% | 5.56% | 21.62% | 7.75% | 148.62% | 0.00% | 17.15% | 141.72% | -inf | 2059.84% |
| ID 11 | -50.27% | -39.01% | -29.51% | -17.99% | -9.89% | 3.81% | -8.02% | 112.22% | -14.64% | 0.00% | 106.33% | -inf | 1743.65% |
| ID 7B | -75.90% | -70.44% | -65.84% | -60.25% | -56.33% | -49.69% | -55.42% | 2.85% | -58.63% | -51.53% | 0.00% | -inf | 793.54% |
| ID 2 | -100.00% | -100.00% | -100.00% | -100.00% | -100.00% | -100.00% | -100.00% | -100.00% | -100.00% | -100.00% | -100.00% | 0.00% | -100.00% |
| ID 5 | -97.30% | -96.69% | -96.18% | -95.55% | -95.11% | -94.37% | -95.01% | -88.49% | -95.37% | -94.58% | -88.81% | -inf | 0.00% |

*Table 10: Comparative matrix of relative mitigation power scores at the scenario group level "car to vulnerable road user", for the US protocol. Each cell shows the relative MPS between vehicle x (row) and vehicle y (column).*

For the mitigation of energy transfer in the event of a C2VRU impact, in the EU protocol ID 6 outperforms ID 9 and ID 1A by over 48% and 43%, respectively (Table 9). ID 4, ID 7B and ID 10, are good performers as well, outperforming most of the other vehicles in the test pool by ~10%.

Again, there are several changes in the relative MPS across EU and US protocols (Table 10), which reflect the differences seen in Tables 7 and 8. Notably, the ID 9 again performs better in the EU protocols, with a predicted ~32% difference in estimated energy transfer from ID 6 on EU roads, versus ~76% difference on US roads.

*Discussion of Comparative Assessment*

We found that the prototype, ID 6, was consistently a strong performer across the scenario groups, FS and MPS metrics, and regional protocols. In addition, in the C2C scenario group, the ID 7B and ID 4 were the closest in terms of performance to the prototype for both regions. The C2VRU scenario group demonstrated more variation between vehicles' performance for the EU and US protocols, with ID 9 performance changing from ~ -30% in the EU to ~ -72% in the US, the ID 4 performance changing from ~ -77% in the EU to ~ -20% in the US, and the ID 7B performance changing from ~ -49% in the EU to ~ -89% in the US. This variation also demonstrates that the frequencies of the individual C2VRU scenarios differ markedly between the two regions.

For the Car-to-Object (C2O) scenario group, ID 6 was the only vehicle to not demonstrate a non-zero performance, both for FS and MPS. Given this, we do not include the relative comparative matrices in the results section. We associate this generalized lack of performance of the pool of tested vehicles to the fact that testing organizations like NCAP do not currently include them in their test protocols. The scenario test results for C2O can be found in Appendix B, Tables 11 - 18.

## Limitations and Considerations

*Considerations in the implementation of the testing protocol*

There are several considerations in the current implementation of the testing protocol, which can impact the scope of our final results and conclusions. For one, in this study, we did not perform multiple iterations of tests, and instead administered one valid test per scenario setting. Given the large number of scenario settings tested for a given scenario and across the protocol, we believe single repetitions of a scenario setting is sufficient to produce the aggregated results presented here. However, due to this implementation, we are unable to test for the significance of our comparisons for a given test, which would require a statistically relevant number of tests to be executed on track. Future implementations could increase the number of repetitions to allow for significance testing in our results, and to further increase the robustness of the study.

In addition, our testing approach involves incrementally increasing vehicle speed for a given collision scenario until the AEB system fails. After failure, we automatically deem the remaining higher speed settings as failed and discontinue further testing for the given scenario. The assumption here is that the complexity of the test is positively correlated with test speed, i.e. a high-speed test is more difficult to pass than a low-speed one. Due to the shorter response times required and the compromised performance of on-vehicle sensors at higher speeds, we believe this assumption to be reasonable. However, future implementations could modify this dynamic approach to both experimentally test this assumption and obtain full results for the protocol.

Finally, human intervention and related potential errors are a natural limitation of track testing, given the large amounts of scenario preparation, measurement tasks, and professional driver decision-making conducted by humans in its implementation. In each test, a professional driver is seated in the tested vehicle who can intervene in the vehicle trajectory if a high-speed crash was deemed imminent by maneuvering towards a safer area of the test track. This intervention was carried out to minimize or avoid damages to the pool of tested vehicles and occurred in approximately 43% of the total tests executed. When such an intervention occurred, impact speed was calculated via speed projections based on vehicle dynamics prior to the intervention. In addition, due to the testing of scenarios not previously tested in other protocols, data collection and some procedural tasks (e.g. regarding the fine alignment of the vehicle at the right offset and speed) were adapted between vehicle tests to improve the implementation and accuracy of the protocol. This may result in slight marginal variations between conditions in the scenario setting across the eight months of the testing campaign.

*Performance of conventional versus LIDAR-based AEB systems*

Our current testing design evaluates the performance of an AEB system as a "black box" containing all components of the system, including the hardware and software. While such results are important for predicting the overall behavior of a vehicle, the strong performance of the ID 6 prototype vehicle—which uses exclusively LIDAR sensors—naturally raises further questions on the behavior (e.g. safety and reliability) of LIDAR-based AEB systems in comparison to more conventional AEB systems which use radar and cameras. Sensor-specific questions are difficult to answer using this "black box" approach, and future study designs could instead separately test and quantify the performance of an AEB system's hardware and software to better isolate each component's performance. However, even so, making general conclusions about the performance of one sensor set versus another within an AEB system is a conceptually challenging task given that hardware and software are designed to work in close conjunction.

*Extrapolating performance of prototype vehicles beyond the test track*

While test track performance is a valuable tool for assessing the capabilities of prototype systems before they reach market, track testing safety performance is not necessarily representative of the final production vehicle to be made available to consumers. While safety is paramount and is the measure we evaluate in this paper, other factors may influence the engineering and tuning of AEB responses prior to consumer release. For example, while a higher rate of *false positives*—in which an AEB activates when not necessary—may not impact and may even improve the safety performance of a vehicle on a test track, false positives can also lead to an uncomfortable or even unsafe driving experience [14] for a consumer. While vehicles available to consumers are designed to have a minimal rate of false positives to ensure a smoother driving experience, the AEB responses of prototype vehicles such as ID 6 may not have already been tuned for driving experience. Consequently, extrapolating a prototype's test track performance to an eventual real-world performance of a production-ready vehicle should be done with caution. To evaluate an AEB's safety performance in relation to the experience it offers drivers, future track testing studies can additionally investigate the rate of false positives or develop additional measures of driving experience.

*Validating realism scores*

Our proposed realism score, based on a computationally tractable probabilistic analysis of behavior models trained on massive driving datasets, represents an important step towards quantifying the realism of generated driving scenarios. However, significant additional research is necessary to fully validate our approach, including (i) a more comprehensive evaluation on a larger set of generated scenarios, (ii) a more systematic human evaluation of the obtained results, and (iii) a rigorous calibration protocol.

*Vehicles homologation vs. geographical context*

In this work we tested vehicles with EU homologation. The performance of such vehicles in the US protocol results from the application of US statistical weights to the respective traffic scenarios. Therefore, we might expect differences in performance of US homologated vehicles when weighed according to US statistical weights. The quantification of the extent of these differences is outside the scope of this work, but future studies could evaluate the impact of homologation on specific country safety metrics.

*The role of ADAS functionality beyond AEB*

Finally, the purpose of this study and the accompanying protocols and scoring frameworks are designed to evaluate the performance of AEB systems specifically. ADAS more broadly consists of several features beyond AEB, including automated steering and forward collision warnings, both of which could supplement a vehicle's performance in pre-collision scenarios in the real-world, had they been activated. As a result, the performance assessed here would not entirely reflect a vehicle's ability in a scenario, given the possibility of additional available ADAS functionality. However, considering that AEB is the primary active ADAS functionality for preventing and mitigating impact in pre-collision scenarios, the results demonstrated in this study should act as strong indicators of a vehicle's performance in a given scenario.

## Concluding Remarks

This manuscript reports on the safety impact assessment and scenarios' realism qualification of a vehicle testing campaign consisting of 12 in-production vehicles and 1 prototype vehicle. The in-production vehicles, which showcase a combination of cameras and radar sensors, were picked for their advanced safety fitment and, thus, expected high safety performance on real roads. The prototype vehicle relies solely on a proprietary LiDAR sensor. The assessment did not single out the sensors performance but focused on both the hardware components and controlling software stacks.

The study makes three important contributions. First, it defines scenarios and scenarios' characteristics that are safety relevant beyond those utilized by well-established testing organizations. Second, it qualifies the degree of realism of such scenarios by utilizing a new metrics introduced in a concurrent paper, thus bringing the stochasticity of real-world events to a finite set of well-defined tests. Third, it introduces a scoring framework to evaluate the probability of vehicles to prevent an accident or mitigate their impact energy transfer in those specific scenarios and compares the performance of 13 vehicles in this framework.

In terms of impact, the study benefits the active safety scientific community in that it provides a complementary approach towards active safety impact assessment to the ones currently experimented by the testing organization, and a new metrics to qualify the realism of the tested scenarios. It also serves well the social mission of insurance companies, in that it underlines the important role of insurance inspired metrics towards understanding and improving road safety, and at the same time their core pricing capabilities since the results of this work (or alike) might be used to build risk scores specific to given car models.

We also ascertained that our test scenario—sub-set of Swiss Re's testing protocol—against which we tested the safety performance of 13 vehicles are realistic. Namely, they are plausible instantiations of driving scenarios that could happen in the real world. However, we highlighted that additional research is needed to fully validate the approach underlying the computation of realism scores. While it might be intuitive to assume that scenarios deemed more realistic are inherently safer, this is not necessarily the case. As described in [13] by Pavone et al., the realism framework focuses on perturbing behavior prediction models to induce challenging situations, potentially leading to collisions. Therefore, a higher realism score, as defined by proximity to the original model's parameter space, might simply imply that the generated scenario reflects a more frequent, yet potentially dangerous, driving behavior. For instance, a scenario where a vehicle makes a slightly wider turn than usual, leading to a sideswipe collision, might be deemed highly realistic due to its frequent occurrence in driving data. However, this doesn't negate the inherent danger of such a maneuver. Therefore, higher realism doesn't necessarily equate to safety but rather points towards a higher likelihood of occurrence, whether safe or unsafe. Future research could explore the nuanced relationship between realism and safety further. At the same time, it is of interest to possibly augment our realism score with additional components derived from human alignment procedures and investigate how our score aligns with other more traditional driving metrics, such as jerk (which is typically used as a proxy for comfort evaluation).

In terms of comparative safety performance of the pool of vehicles, there are noteworthy differences between them, both in terms of capability to prevent and mitigate accidents. The prototype vehicle outperforms, on average, the in-production vehicles by at least ~27% and ~20% in terms of capability to prevent accidents in the EU and US setting, respectively. The average value was computed by accounting for the smallest comparative improvements in car-to-car and car-to-vulnerable road users' performance. The prototype was the only vehicle to exhibit a non-zero performance in car-to-objects scenarios. In terms of capability to mitigate an accident, again the prototype vehicle outperforms the in-production vehicles by, on average, at least ~34% and ~32% in terms of capability to prevent accidents in the EU and US setting, respectively. These results should be contextualized according to the limitations mentioned in the *Limitations and Considerations* section.

Future work should be devoted to understanding the extent to which the performance of the prototype vehicle can be exploited in the real world, namely whether such strong behavior goes to the expense of drivability (e.g., appearance of false positives). Further, time should be spent to identify and formalize the relationship, if any, between comfort, realism, and safety. We think that this for-now-unknown relationship could foster the understanding of whether an increased level of automation, which is typically associated with higher safety, is indeed closer or further from the behavioral model of a generalized human driver.

## Acknowledgments

**Giovanni Vadala**, Product and Market Strategy at Swiss Reinsurance Company has been instrumental in improving data visualization, results presentation, and counter-checking overall text consistency.

**Gustavo Imbrizi Prado**, Vehicle Safety Engineer Expert at Swiss Reinsurance Company has been instrumental in validating the underlying dataset, challenging the testing results, improving data accuracy, and supporting data visualization.

**Hans-Georg Hermann**, Senior Product Manager Vehicle Safety & AV at Swiss Reinsurance Company has been of great help in challenging and evaluating the strategic impact of the work and provide valuable support in that respect.

# Appendix A

In this section we will define the acronyms and provide a schema of the scenarios of our test protocol.

| Acronym | Diagram | Description |
|---|---|---|
| CCRs | 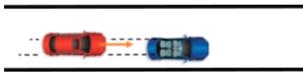 | Car-to-car rear end collision in which the vehicle in the front is stopped |
| CCRm | 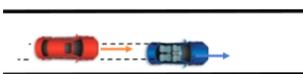 | Car-to-car rear end collision in which the vehicle in the front is moving at a lower speed |
| CCFtap | 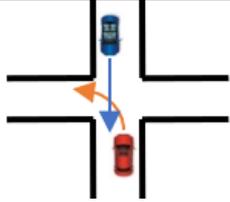 | Front Turn across path: one vehicle is turning left, crashing into a vehicle coming from the opposite direction |
| CCscp Left | 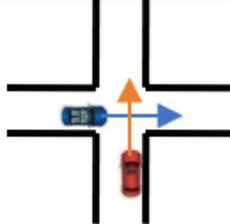 | Straight crossing path from the left: one vehicle is going straight, crossing an intersection, and it crashes into a vehicle coming from the left |
| CCscp Right | 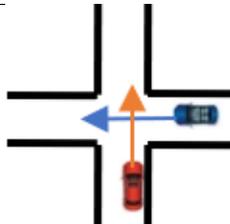 | Straight crossing path from the right: one vehicle is going straight, crossing an intersection, and it crashes into a vehicle coming from the right |
| CPFA | 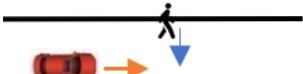 | Car to Pedestrian Adult Nearside: the pedestrian is crossing the road, coming from the left of the car |
| CPNA | 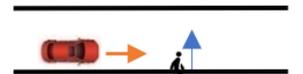 | Car to Pedestrian Adult Nearside: the pedestrian is crossing the road, coming from the right of the car |
| CPNAO | 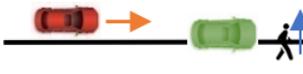 | Car to Pedestrian Adult Nearside obstructed: variant of the CPNA scenario in which the view is obstructed by an obstacle |
| CPNCO | 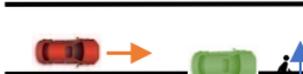 | Car to Pedestrian Child Nearside obstructed: variant of the CPNAo scenario in which the pedestrian is a child |
| CPLAs | 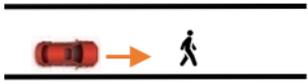 | Car to Pedestrian Adult Stationary: the pedestrian is stationary in the middle of the road |
| CPLA | 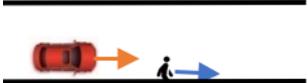 | Car to Pedestrian Adult Longitudinal: the car and the pedestrian are going in the same direction |
| CPTA Farside o | 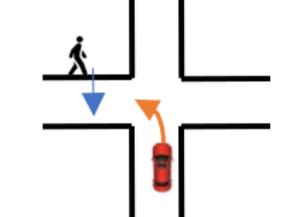 | Car to Pedestrian Turning Adult Farside opposite direction: the vehicle is turning left, crashing into a pedestrian that is crossing the road, coming from the right of the car |
| CPTA Farside s | 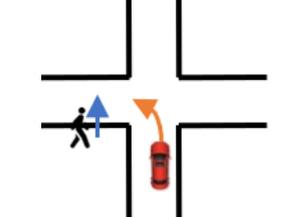 | Car to Pedestrian Turning Adult Farside same direction: the vehicle is turning left, crashing into a pedestrian that is crossing the road, coming from the left of the car |
| CPTA Nearside s | 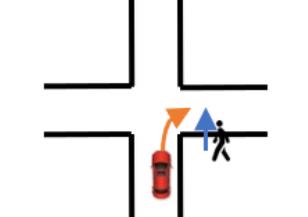 | Car to Pedestrian Turning Adult Nearside same direction: the vehicle is turning right, crashing into a pedestrian that is crossing the road, coming from the right of the car |
| CPTA Nearside o | 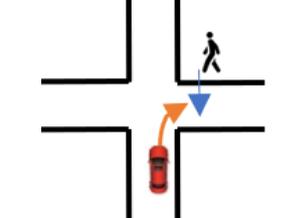 | Car to Pedestrian Turning Adult Nearside opposite direction: the vehicle is turning right, crashing into a pedestrian that is crossing the road, coming from the left of the car |
| CPTC Farside o | 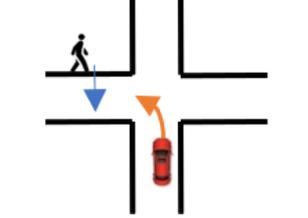 | Car to Pedestrian Turning Child Farside opposite direction: the vehicle is turning left, crashing into a pedestrian that is crossing the road, coming from the right of the car |
| CPTC Farside s | 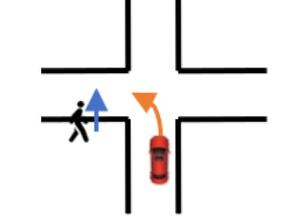 | Car to Pedestrian Turning Child Farside same direction: the vehicle is turning left, crashing into a pedestrian that is crossing the road, coming from the left of the car |
| CPTC Nearside s | 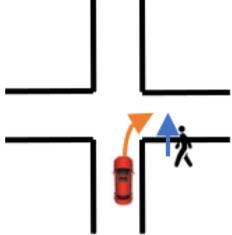 | Car to Pedestrian Turning Child Nearside same direction: the vehicle is turning right, crashing into a pedestrian that is crossing the road, coming from the right of the car |
| CPTC Nearside o | 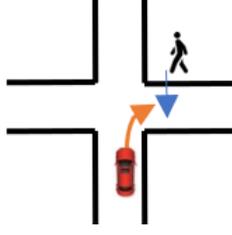 | Car to Pedestrian Turning A Child dult Nearside opposite direction: the vehicle is turning right, crashing into a pedestrian that is crossing the road, coming from the left of the car |
| CBFA | 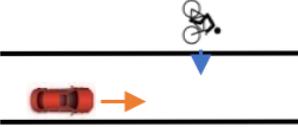 | Car to Bicycle Adult Nearside: the cyclist is crossing the road, coming from the left of the car |
| CBNA | 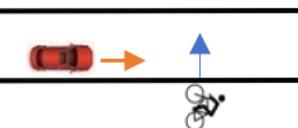 | Car to Bicycle Adult Nearside: the cyclist is crossing the road, coming from the right of the car |
| CBNAO | 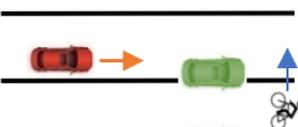 | Car to Bicycle Nearside Crossing, Obstructed: variant of the CBNA scenario in which the view is obstructed by an obstacle |
| CBLA | 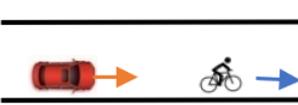 | Car to Bicycle Longitudinal: the car and the bicycle are going in the same direction |
| CBTA Farside | 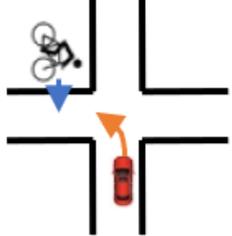 | Car to Bicycle Turning Farside: the vehicle is turning left, crashing into a cyclist that is crossing the road, coming from the right of the car |
| CBTA Nearside | 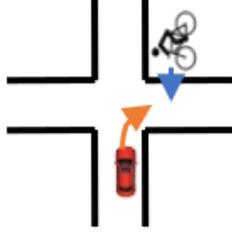 | Car to Bicycle Turning Adult Nearside opposite direction: the vehicle is turning right, crashing into a cyclist that is crossing the road, coming from the left of the car |
| Scooter | 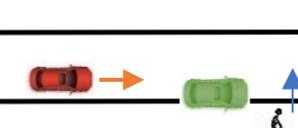 | Car to E-Scooter Crossing Nearside, Obstructed: the scooter is crossing the road, coming from the right of the car obstructed by another vehicle |
| Bobbycar | 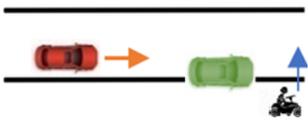 | Car to Bobby Car Crossing Nearside obstructed: the bobbycar is crossing the road, coming from the right of the car obstructed by another vehicle |
| PALLETS | 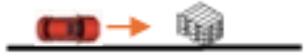 | Collision between a car and three stacked pallets. |
| TIRE | 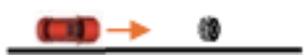 | Collision between a car and a tire |
| CPNCO Group | 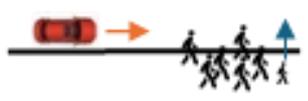 | Car to Pedestrian Child Nearside obstructed: the child is crossing the road, coming from the right of the car obstructed by a group of people |

**Appendix B**

In Tables 11 - 18, the performance of the vehicles is reported at scenario level, the highest level of granularity discussed in this manuscript. By highest level of granularity, we mean that a value of FS and MPS is provided for each scenario of table 2, both in day and night conditions.

Given the experimental nature of some of the tests performed in this study, and the limitations discussed in the *Limitations and Considerations* section, we introduced a measure of uncertainty in the computation of the scores that account for the error in the measurements of the tests speed, which is the most characterizing variable of the test protocol. This error is obtained by computing an upper and lower limit of the FS and MPS scores by assuming the speed at which the vehicle fails to be 5 km/h lower/higher than the nominal one defined in the testing protocol.
The values in each cell (x, y), where x is the scenario level and y the VUT, are therefore shown in terms of mean value and standard deviation. For CCRs and CCRm scenarios, which have been tested in more than one overlap (see Table 2), we have aggregated the performance of each single tested overlap.

For the sake of interpretation of Tables 11 - 18, consider Table 11, which shows a comparison in terms of FS performance using EU statistical weights in daylight conditions, and the scenario CCRm.
- A higher FS is indicative of a stronger performance, namely a higher capability to prevent CCRm accidents. The ID 1A and 1B vehicles, the ID 6, the ID 7B, the ID 9, and the ID 11 are similarly top performers (mean + standard deviation), whilst the ID 2 is the worst performer.
- ID 9 and the ID 11 are the only two vehicles of the tested pool that show a non-zero performance in the CBNAO scenario, day light conditions, whereas the prototype, ID 6, is the only vehicle showing a non-zero performance in car-to-object scenarios (Pallets and Tire), both in day light and night conditions.

A similar reading applies to the tables reporting mitigation power scores. A high MPS in each scenario is indicative of strong mitigation power capabilities in that specific scenario.

| | FREQ_SCORE_MEAN_DAY_EU | | | | | | | | | | | | |
|---|---|---|---|---|---|---|---|---|---|---|---|---|---|
| MODEL | ID 1A | ID 1B | ID 2 | ID 3 | ID 4 | ID 5 | ID 6 | ID 7A | ID 7B | ID 8 | ID 9 | ID 10 | ID 11 |
| CBTA Farside | 0±0 | 0±0 | 0±0 | 0±0 | 0.63±0.05 | 0±0 | 0.81±0.05 | 0.63±0.05 | 0.36±0.06 | 0±0 | 0.36±0.06 | 0±0 | 0±0 |
| CBNA | 0.74±0.02 | 0±0 | 0±0 | 0±0 | 0±0 | 0±0 | 0.74±0.02 | 0.65±0.02 | 0.74±0.02 | 0±0 | 0.7±0.01 | 0.72±0.02 | 0±0 |
| CBTA Nearside | 0.36±0.06 | 0±0 | 0±0 | 0±0 | 0±0 | 0±0 | 0.78±0.05 | 0±0 | 0±0 | 0±0 | 0.36±0.06 | 0±0 | 0±0 |
| CBNAO | 0±0 | 0±0 | 0±0 | 0±0 | 0±0 | 0±0 | 0±0 | 0±0 | 0±0 | 0±0 | 0.65±0.02 | 0±0 | 0.7±0.01 |
| CCRm | 0.66±0.13 | 0.66±0.13 | 0.46±0.1 | 0±0 | 0.54±0.12 | 0.46±0.1 | 0.67±0.12 | 0.62±0.13 | 0.68±0.11 | 0.59±0.13 | 0.67±0.12 | 0.62±0.13 | 0.66±0.13 |
| CCRs | 0.86±0.05 | 0.81±0.05 | 0±0 | 0±0 | 0.69±0.01 | 0±0 | 0.9±0.07 | 0.76±0.04 | 0.89±0.06 | 0.63±0.01 | 0.85±0.06 | 0.82±0.05 | 0.81±0.04 |
| CPTA Farside s | 0.63±0.05 | 0±0 | 0±0 | 0±0 | 0.63±0.05 | 0±0 | 0.81±0.05 | 0±0 | 0±0 | 0±0 | 0.81±0.05 | 0±0 | 0±0 |
| CPTA Nearside o | 0.16±0.05 | 0±0 | 0±0 | 0±0 | 0.16±0.05 | 0±0 | 0.67±0.14 | 0±0 | 0±0 | 0.31±0.02 | 0.31±0.02 | 0±0 | 0±0 |
| CPLA | 0.71±0.03 | 0.81±0.03 | 0±0 | 0.59±0.02 | 0.81±0.03 | 0±0 | 0.91±0.02 | 0.71±0 | 0.91±0.02 | 0.59±0.02 | 0.91±0.02 | 0.71±0.03 | 0.71±0 |
| CPNAO | 0±0 | 0±0 | 0±0 | 0±0 | 0±0 | 0±0 | 0.78±0.15 | 0±0 | 0.64±0.11 | 0.64±0.11 | 0.73±0.14 | 0.68±0.11 | 0±0 |
| CPTA Nearside s | 0.67±0.14 | 0±0 | 0±0 | 0±0 | 0±0 | 0±0 | 0.67±0.14 | 0±0 | 0±0 | 0±0 | 0.31±0.02 | 0±0 | 0±0 |
| CPLAs | NA | NA | NA | NA | NA | NA | NA | NA | NA | NA | NA | NA | NA |
| CPTC Farside s | 0.81±0.05 | 0±0 | 0±0 | 0±0 | 0.63±0.05 | 0±0 | 0.81±0.05 | 0±0 | 0.36±0.06 | 0.63±0.05 | 0.36±0.06 | 0±0 | 0±0 |
| CPNCO | 0.62±0.08 | 0±0 | 0±0 | 0±0 | 0±0 | 0±0 | 0.72±0.14 | 0±0 | 0.64±0.11 | 0.64±0.11 | 0.64±0.11 | 0±0 | 0±0 |
| CPNCO group | 0.78±0.15 | 0.72±0.12 | 0±0 | 0±0 | 0±0 | 0±0 | 0.82±0.15 | 0±0 | 0.72±0.12 | 0.78±0.15 | 0.74±0.13 | 0.62±0.08 | 0±0 |
| CPTC Nearside s | 0.31±0.02 | 0±0 | 0±0 | 0±0 | 0±0 | 0±0 | 0.67±0.14 | 0±0 | 0±0 | 0±0 | 0.31±0.02 | 0±0 | 0±0 |
| CCFtap | 0.42±0.06 | 0±0 | 0±0 | 0±0 | 0.92±0.02 | 0±0 | 0.92±0.02 | 0.42±0.06 | 0.52±0.05 | 0.79±0.05 | 0.45±0.08 | 0±0 | 0.52±0.05 |
| CCscp left | 0±0 | 0±0 | 0±0 | 0±0 | 0.33±0.1 | 0±0 | 0.4±0.1 | 0±0 | 0.61±0.09 | 0±0 | 0±0 | 0±0 | 0±0 |
| CCscp right | 0±0 | 0±0 | 0±0 | 0±0 | 0±0 | 0±0 | 0.87±0.03 | 0±0 | 0.56±0.1 | 0±0 | 0±0 | 0±0 | 0±0 |
| Pallets | 0±0 | 0±0 | 0±0 | 0±0 | 0±0 | 0±0 | 0.93±0.02 | 0±0 | 0±0 | 0±0 | 0±0 | 0±0 | 0±0 |
| Tire | 0±0 | 0±0 | 0±0 | 0±0 | 0±0 | 0±0 | 0.19±0.05 | 0±0 | 0±0 | 0±0 | 0±0 | 0±0 | 0±0 |

Table 11: Comparative assessment of the FS performance mean ± standard deviation of each vehicle of the test pool for the full set of scenarios of the test protocol, using EU statistical weights, in day light conditions.

| | FREQ_SCORE_MEAN_DAY_US | | | | | | | | | | | | |
|---|---|---|---|---|---|---|---|---|---|---|---|---|---|
| MODEL | ID 1A | ID 1B | ID 2 | ID 3 | ID 4 | ID 5 | ID 6 | ID 7A | ID 7B | ID 8 | ID 9 | ID 10 | ID 11 |
| CBTA Farside | 0±0 | 0±0 | 0±0 | 0±0 | 0.62±0.04 | 0±0 | 0.77±0.01 | 0.62±0.04 | 0.38±0.08 | 0±0 | 0.38±0.08 | 0±0 | 0±0 |
| CBNA | 0.73±0.01 | 0±0 | 0±0 | 0±0 | 0±0 | 0±0 | 0.73±0.01 | 0.63±0 | 0.73±0.01 | 0±0 | 0.69±0 | 0.71±0.01 | 0±0 |
| CBTA Nearside | 0.38±0.08 | 0±0 | 0±0 | 0±0 | 0±0 | 0±0 | 0.75±0.02 | 0±0 | 0±0 | 0±0 | 0.38±0.08 | 0±0 | 0±0 |
| CBNAO | 0±0 | 0±0 | 0±0 | 0±0 | 0±0 | 0±0 | 0±0 | 0±0 | 0±0 | 0±0 | 0.63±0 | 0±0 | 0.69±0 |
| CCRm | 0.79±0 | 0.79±0 | 0.59±0.03 | 0±0 | 0.68±0.02 | 0.59±0.03 | 0.8±0 | 0.76±0.01 | 0.8±0 | 0.73±0.01 | 0.8±0 | 0.76±0.01 | 0.79±0 |
| CCRs | 0.8±0 | 0.78±0.01 | 0±0 | 0±0 | 0.69±0 | 0±0 | 0.84±0 | 0.74±0.01 | 0.83±0 | 0.69±0.01 | 0.8±0.01 | 0.78±0 | 0.78±0 |
| CPTA Farside s | 0.62±0.04 | 0±0 | 0±0 | 0±0 | 0.62±0.04 | 0±0 | 0.77±0.01 | 0±0 | 0±0 | 0±0 | 0.77±0.01 | 0±0 | 0±0 |
| CPTA Nearside o | 0.21±0 | 0±0 | 0±0 | 0±0 | 0.21±0 | 0±0 | 0.54±0.01 | 0±0 | 0±0 | 0.34±0.05 | 0.34±0.05 | 0±0 | 0±0 |
| CPLA | 0.71±0.02 | 0.8±0.02 | 0±0 | 0.6±0.03 | 0.8±0.02 | 0±0 | 0.89±0.01 | 0.71±0 | 0.89±0.01 | 0.6±0.03 | 0.89±0.01 | 0.71±0.02 | 0.71±0 |
| CPNAO | 0±0 | 0±0 | 0±0 | 0±0 | 0±0 | 0±0 | 0.64±0.01 | 0±0 | 0.54±0.02 | 0.54±0.02 | 0.59±0.01 | 0.57±0 | 0±0 |
| CPTA Nearside s | 0.54±0.01 | 0±0 | 0±0 | 0±0 | 0±0 | 0±0 | 0.54±0.01 | 0±0 | 0±0 | 0±0 | 0.34±0.05 | 0±0 | 0±0 |
| CPLAs | NA | NA | NA | NA | NA | NA | NA | NA | NA | NA | NA | NA | NA |
| CPTC Farside s | 0.77±0.01 | 0±0 | 0±0 | 0±0 | 0.62±0.04 | 0±0 | 0.77±0.01 | 0±0 | 0.38±0.08 | 0.62±0.04 | 0.38±0.08 | 0±0 | 0±0 |
| CPNCO | 0.54±0 | 0±0 | 0±0 | 0±0 | 0±0 | 0±0 | 0.6±0.01 | 0±0 | 0.54±0.02 | 0.54±0.02 | 0.54±0.02 | 0±0 | 0±0 |
| CPNCO group | 0.64±0.01 | 0.6±0 | 0±0 | 0±0 | 0±0 | 0±0 | 0.67±0.01 | 0±0 | 0.6±0 | 0.64±0.01 | 0.61±0 | 0.54±0 | 0±0 |
| CPTC Nearside s | 0.34±0.05 | 0±0 | 0±0 | 0±0 | 0±0 | 0±0 | 0.54±0.01 | 0±0 | 0±0 | 0±0 | 0.34±0.05 | 0±0 | 0±0 |
| CCFtap | 0.4±0.04 | 0±0 | 0±0 | 0±0 | 0.91±0.01 | 0±0 | 0.91±0.01 | 0.4±0.04 | 0.5±0.04 | 0.77±0.03 | 0.43±0.06 | 0±0 | 0.5±0.04 |
| CCscp left | 0±0 | 0±0 | 0±0 | 0±0 | 0.26±0.03 | 0±0 | 0.33±0.02 | 0±0 | 0.54±0.02 | 0±0 | 0±0 | 0±0 | 0±0 |
| CCscp right | 0±0 | 0±0 | 0±0 | 0±0 | 0±0 | 0±0 | 0.84±0.01 | 0±0 | 0.49±0.03 | 0±0 | 0±0 | 0±0 | 0±0 |
| Pallets | 0±0 | 0±0 | 0±0 | 0±0 | 0±0 | 0±0 | 0.93±0.02 | 0±0 | 0±0 | 0±0 | 0±0 | 0±0 | 0±0 |
| Tire | 0±0 | 0±0 | 0±0 | 0±0 | 0±0 | 0±0 | 0.17±0.03 | 0±0 | 0±0 | 0±0 | 0±0 | 0±0 | 0±0 |

Table 12: Comparative assessment of the FS performance mean ± standard deviation of each vehicle of the test pool for the full set of scenarios of the test protocol, using US statistical weights, in day light conditions.

| | FREQ_SCORE_MEAN_NIGHT_EU | | | | | | | | | | | | |
|---|---|---|---|---|---|---|---|---|---|---|---|---|---|
| MODEL | ID 1A | ID 1B | ID 2 | ID 3 | ID 4 | ID 5 | ID 6 | ID 7A | ID 7B | ID 8 | ID 9 | ID 10 | ID 11 |
| CBTA Farside | NA | NA | NA | NA | NA | NA | NA | NA | NA | NA | NA | NA | NA |
| CBNA | NA | NA | NA | NA | NA | NA | NA | NA | NA | NA | NA | NA | NA |
| CBTA Nearside | NA | NA | NA | NA | NA | NA | NA | NA | NA | NA | NA | NA | NA |
| CBNAO | NA | NA | NA | NA | NA | NA | NA | NA | NA | NA | NA | NA | NA |
| CCRm | 0.67±0.12 | 0.64±0.14 | 0±0 | 0±0 | 0.54±0.14 | 0±0 | 0.67±0.12 | 0.62±0.14 | 0.62±0.14 | 0.54±0.14 | 0.66±0.13 | 0.54±0.14 | 0.62±0.14 |
| CCRs | 0.86±0.05 | 0±0 | 0±0 | 0±0 | 0±0 | 0±0 | 0.9±0.07 | 0±0 | 0.89±0.06 | 0.68±0.01 | 0.82±0.04 | 0.68±0.01 | 0.68±0.01 |
| CPTA Farside s | 0.36±0.02 | 0±0 | 0±0 | 0±0 | 0±0 | 0±0 | 0.81±0.05 | 0±0 | 0±0 | 0±0 | 0.81±0.05 | 0±0 | 0±0 |
| CPTA Nearside o | NA | NA | NA | NA | NA | NA | NA | NA | NA | NA | NA | NA | NA |
| CPLA | 0.58±0.02 | 0.71±0 | 0±0 | 0.58±0.02 | 0.81±0.03 | 0±0 | 0.91±0.02 | 0.81±0.03 | 0±0 | 0.58±0.02 | 0±0 | 0.58±0.02 | 0.59±0.01 |
| CPNAO | 0±0 | 0±0 | 0±0 | 0±0 | 0±0 | 0±0 | 0.78±0.02 | 0±0 | 0±0 | 0±0 | 0.72±0.12 | 0±0 | 0±0 |
| CPTA Nearside s | 0.67±0.14 | 0±0 | 0±0 | 0±0 | 0±0 | 0±0 | 0.67±0.14 | 0±0 | 0±0 | 0±0 | 0.31±0.02 | 0±0 | 0±0 |
| CPLAs | 0.75±0.05 | 0±0 | 0±0 | 0.63±0.02 | 0.82±0.06 | 0±0 | 0.9±0.07 | 0.63±0.02 | 0±0 | 0.63±0.02 | 0±0 | 0.63±0.02 | 0±0 |
| CPTC Farside s | NA | NA | NA | NA | NA | NA | NA | NA | NA | NA | NA | NA | NA |
| CPNCO | 0.62±0.08 | 0±0 | 0±0 | 0±0 | 0±0 | 0±0 | 0.82±0.15 | 0±0 | 0±0 | 0±0 | 0±0 | 0±0 | 0±0 |
| CPNCO group | NA | NA | NA | NA | NA | NA | NA | NA | NA | NA | NA | NA | NA |
| CPTC Nearside s | NA | NA | NA | NA | NA | NA | NA | NA | NA | NA | NA | NA | NA |
| CCFtap | 0.32±0.02 | 0±0 | 0±0 | 0±0 | 0.47±0.06 | 0±0 | 0.93±0.01 | 0±0 | 0.32±0.02 | 0±0 | 0.45±0.04 | 0±0 | 0±0 |
| CCscp left | 0±0 | 0±0 | 0±0 | 0±0 | 0.51±0.15 | 0±0 | 0.57±0.15 | 0±0 | 0.69±0.08 | 0±0 | 0±0 | 0±0 | 0±0 |
| CCscp right | 0±0 | 0±0 | 0±0 | 0±0 | 0±0 | 0±0 | 0±0 | 0±0 | 0±0 | 0.51±0.1 | 0±0 | 0±0 | 0±0 |
| Pallets | 0±0 | 0±0 | 0±0 | 0±0 | 0±0 | 0±0 | 0.93±0.02 | 0±0 | 0±0 | 0±0 | 0±0 | 0±0 | 0±0 |
| Tire | 0±0 | 0±0 | 0±0 | 0±0 | 0±0 | 0±0 | 0.81±0.05 | 0±0 | 0±0 | 0±0 | 0±0 | 0±0 | 0±0 |

Table 13: Comparative assessment of the FS performance mean ± standard deviation of each vehicle of the test pool for the full set of scenarios of the test protocol, using EU statistical weights, in night conditions.

| | FREQ_SCORE_MEAN_NIGHT_US | | | | | | | | | | | | |
|---|---|---|---|---|---|---|---|---|---|---|---|---|---|
| MODEL | ID 1A | ID 1B | ID 2 | ID 3 | ID 4 | ID 5 | ID 6 | ID 7A | ID 7B | ID 8 | ID 9 | ID 10 | ID 11 |
| CBTA Farside | NA | NA | NA | NA | NA | NA | NA | NA | NA | NA | NA | NA | NA |
| CBNA | NA | NA | NA | NA | NA | NA | NA | NA | NA | NA | NA | NA | NA |
| CBTA Nearside | NA | NA | NA | NA | NA | NA | NA | NA | NA | NA | NA | NA | NA |
| CBNAO | NA | NA | NA | NA | NA | NA | NA | NA | NA | NA | NA | NA | NA |
| CCRm | 0.8±0 | 0.78±0 | 0±0 | 0±0 | 0.68±0 | 0±0 | 0.8±0 | 0.76±0 | 0.76±0 | 0.68±0 | 0.79±0 | 0.68±0 | 0.76±0 |
| CCRs | 0.8±0 | 0±0 | 0±0 | 0±0 | 0±0 | 0±0 | 0.84±0 | 0±0 | 0.83±0 | 0.69±0 | 0.78±0 | 0.69±0 | 0.69±0 |
| CPTA Farside s | 0.38±0 | 0±0 | 0±0 | 0±0 | 0±0 | 0±0 | 0.77±0.01 | 0±0 | 0±0 | 0±0 | 0.77±0.01 | 0±0 | 0±0 |
| CPTA Nearside o | NA | NA | NA | NA | NA | NA | NA | NA | NA | NA | NA | NA | NA |
| CPLA | 0.6±0 | 0.71±0 | 0±0 | 0.6±0 | 0.8±0.02 | 0±0 | 0.89±0.01 | 0.8±0.02 | 0±0 | 0.6±0 | 0±0 | 0.6±0 | 0.6±0 |
| CPNAO | 0±0 | 0±0 | 0±0 | 0±0 | 0±0 | 0±0 | 0.64±0.01 | 0±0 | 0±0 | 0±0 | 0.6±0 | 0±0 | 0±0 |
| CPTA Nearside s | 0.54±0.01 | 0±0 | 0±0 | 0±0 | 0±0 | 0±0 | 0.54±0.01 | 0±0 | 0±0 | 0±0 | 0.34±0.05 | 0±0 | 0±0 |
| CPLAs | 0.7±0 | 0±0 | 0±0 | 0.61±0 | 0.76±0 | 0±0 | 0.84±0.01 | 0.61±0 | 0±0 | 0.61±0 | 0±0 | 0.61±0 | 0±0 |
| CPTC Farside s | NA | NA | NA | NA | NA | NA | NA | NA | NA | NA | NA | NA | NA |
| CPNCO | 0.54±0 | 0±0 | 0±0 | 0±0 | 0±0 | 0±0 | 0.67±0.01 | 0±0 | 0±0 | 0±0 | 0±0 | 0±0 | 0±0 |
| CPNCO group | NA | NA | NA | NA | NA | NA | NA | NA | NA | NA | NA | NA | NA |
| CPTC Nearside s | NA | NA | NA | NA | NA | NA | NA | NA | NA | NA | NA | NA | NA |
| CCFtap | 0.3±0 | 0±0 | 0±0 | 0±0 | 0.46±0.04 | 0±0 | 0.93±0.01 | 0±0 | 0.3±0 | 0±0 | 0.44±0.02 | 0±0 | 0±0 |
| CCscp left | 0±0 | 0±0 | 0±0 | 0±0 | 0.44±0.07 | 0±0 | 0.5±0.07 | 0±0 | 0.63±0.02 | 0±0 | 0±0 | 0±0 | 0±0 |
| CCscp right | 0±0 | 0±0 | 0±0 | 0±0 | 0±0 | 0±0 | 0±0 | 0±0 | 0±0 | 0.43±0.02 | 0±0 | 0±0 | 0±0 |
| Pallets | 0±0 | 0±0 | 0±0 | 0±0 | 0±0 | 0±0 | 0.93±0.01 | 0±0 | 0±0 | 0±0 | 0±0 | 0±0 | 0±0 |
| Tire | 0±0 | 0±0 | 0±0 | 0±0 | 0±0 | 0±0 | 0.79±0.03 | 0±0 | 0±0 | 0±0 | 0±0 | 0±0 | 0±0 |

Table 14: Comparative assessment of the FS performance mean ± standard deviation of each vehicle of the test pool for the full set of scenarios of the test protocol, using US statistical weights, in night conditions.

| | MIT_POW_DAY_EU | | | | | | | | | | | | |
|---|---|---|---|---|---|---|---|---|---|---|---|---|---|
| MODEL | ID 1A | ID 1B | ID 2 | ID 3 | ID 4 | ID 5 | ID 6 | ID 7A | ID 7B | ID 8 | ID 9 | ID 10 | ID 11 |
| CBTA Farside | 0.25±0.2 | 0±0 | 0±0 | 0±0 | 0.57±0.05 | 0±0 | 0.87±0.03 | 0.57±0.05 | 0.26±0.05 | 0±0 | 0.26±0.05 | 0±0 | 0±0 |
| CBNA | 0.78±0.07 | 0.48±0.13 | 0±0 | 0±0 | 0.72±0.08 | 0±0 | 0.79±0.07 | 0.69±0.08 | 0.79±0.07 | 0.52±0.12 | 0.79±0.07 | 0.74±0.08 | 0±0 |
| CBTA Nearside | 0.34±0.09 | 0±0 | 0±0 | 0±0 | 0±0 | 0±0 | 0.82±0.03 | 0±0 | 0±0 | 0±0 | 0.26±0.05 | 0±0 | 0±0 |
| CBNAO | 0.51±0.05 | 0±0 | 0±0 | 0±0 | 0.37±0.14 | 0±0 | 0.55±0.05 | 0.51±0.05 | 0.46±0.05 | 0±0 | 0.68±0.05 | 0.55±0.05 | 0.68±0.05 |
| CCRm | 0.64±0.16 | 0.73±0.15 | 0.23±0.07 | 0±0 | 0.48±0.12 | 0.4±0.12 | 0.75±0.14 | 0.62±0.15 | 0.78±0.11 | 0.55±0.14 | 0.77±0.14 | 0.63±0.15 | 0.78±0.13 |
| CCRs | 0.6±0.09 | 0.66±0.04 | 0±0 | 0±0 | 0.58±0.05 | 0.51±0.03 | 0.9±0.02 | 0.56±0.05 | 0.77±0.03 | 0.47±0.04 | 0.71±0.04 | 0.65±0.04 | 0.73±0.07 |
| CPTA Farside s | 0.64±0.12 | 0±0 | 0±0 | 0±0 | 0.57±0.05 | 0±0 | 0.87±0.03 | 0±0 | 0±0 | 0±0 | 0.87±0.03 | 0±0 | 0±0 |
| CPTA Nearside o | 0.16±0.06 | 0±0 | 0±0 | 0±0 | 0.31±0.04 | 0±0 | 0.74±0.04 | 0.1±0.06 | 0±0 | 0.24±0.04 | 0.24±0.04 | 0±0 | 0±0 |
| CPLA | 0.68±0.08 | 0.78±0.06 | 0±0 | 0.51±0.07 | 0.79±0.06 | 0.21±0.05 | 0.88±0.01 | 0.68±0.01 | 0.88±0.01 | 0.43±0.02 | 0.88±0.01 | 0.65±0.07 | 0.78±0.06 |
| CPNAO | 0.27±0.19 | 0.4±0.13 | 0±0 | 0.39±0.13 | 0.57±0.17 | 0.45±0.14 | 0.77±0.17 | 0.42±0.13 | 0.59±0.17 | 0.51±0.12 | 0.7±0.17 | 0.77±0.17 | 0.47±0.14 |
| CPTA Nearside s | 0.74±0.04 | 0±0 | 0±0 | 0±0 | 0±0 | 0±0 | 0.74±0.04 | 0±0 | 0±0 | 0±0 | 0.24±0.04 | 0±0 | 0±0 |
| CPLAs | NA | NA | NA | NA | NA | NA | NA | NA | NA | NA | NA | NA | NA |
| CPTC Farside s | 0.87±0.03 | 0±0 | 0±0 | 0±0 | 0.57±0.05 | 0±0 | 0.87±0.03 | 0±0 | 0.39±0.14 | 0.6±0.06 | 0.45±0.15 | 0±0 | 0±0 |
| CPNCO | 0.62±0.14 | 0.49±0.11 | 0±0 | 0±0 | 0.57±0.17 | 0.51±0.12 | 0.7±0.18 | 0.39±0.13 | 0.6±0.17 | 0.51±0.12 | 0.59±0.17 | 0.44±0.14 | 0±0 |
| CPNCO group | 0.77±0.17 | 0.75±0.16 | 0±0 | 0.37±0.13 | 0.73±0.15 | 0.56±0.17 | 0.82±0.16 | 0.51±0.12 | 0.73±0.17 | 0.76±0.18 | 0.81±0.16 | 0.71±0.14 | 0±0 |
| CPTC Nearside s | 0.33±0.05 | 0±0 | 0±0 | 0±0 | 0±0 | 0±0 | 0.74±0.04 | 0±0 | 0±0 | 0±0 | 0.24±0.04 | 0±0 | 0±0 |
| CCFtap | 0.27±0.08 | 0±0 | 0±0 | 0±0 | 0.84±0.05 | 0±0 | 0.83±0.05 | 0.35±0.08 | 0.44±0.05 | 0.68±0.06 | 0.44±0.07 | 0±0 | 0.62±0.05 |
| CCscp left | 0.17±0.02 | 0±0 | 0±0 | 0±0 | 0.39±0.06 | 0±0 | 0.45±0.07 | 0±0 | 0.51±0.09 | 0±0 | 0.17±0.03 | 0±0 | 0±0 |
| CCscp right | 0.15±0.02 | 0±0 | 0±0 | 0±0 | 0±0 | 0±0 | 0.83±0.04 | 0.19±0.03 | 0.45±0.06 | 0±0 | 0.19±0.03 | 0±0 | 0±0 |
| Pallets | 0±0 | 0±0 | 0±0 | 0±0 | 0±0 | 0±0 | 0.9±0.02 | 0±0 | 0±0 | 0±0 | 0±0 | 0±0 | 0±0 |
| Tire | 0±0 | 0±0 | 0±0 | 0±0 | 0±0 | 0±0 | 0.21±0.1 | 0±0 | 0±0 | 0±0 | 0±0 | 0±0 | 0±0 |

Table 15: Comparative assessment of the MPS performance mean ± standard deviation of each vehicle of the test pool for the full set of scenarios of the test protocol, using EU statistical weights, in day light conditions.

| | MIT_POW_DAY_US | | | | | | | | | | | | |
|---|---|---|---|---|---|---|---|---|---|---|---|---|---|
| MODEL | ID 1A | ID 1B | ID 2 | ID 3 | ID 4 | ID 5 | ID 6 | ID 7A | ID 7B | ID 8 | ID 9 | ID 10 | ID 11 |
| CBTA Farside | 0.27±0.22 | 0±0 | 0±0 | 0±0 | 0.58±0.06 | 0±0 | 0.86±0.02 | 0.58±0.06 | 0.29±0.08 | 0±0 | 0.29±0.08 | 0±0 | 0±0 |
| CBNA | 0.75±0.03 | 0.43±0.08 | 0±0 | 0±0 | 0.69±0.05 | 0±0 | 0.75±0.03 | 0.65±0.03 | 0.75±0.03 | 0.47±0.07 | 0.75±0.03 | 0.69±0.02 | 0±0 |
| CBTA Nearside | 0.37±0.12 | 0±0 | 0±0 | 0±0 | 0±0 | 0±0 | 0.82±0.03 | 0±0 | 0±0 | 0±0 | 0.29±0.08 | 0±0 | 0±0 |
| CBNAO | 0.46±0 | 0±0 | 0±0 | 0±0 | 0.34±0.17 | 0±0 | 0.5±0 | 0.46±0 | 0.42±0 | 0±0 | 0.63±0 | 0.5±0 | 0.63±0 |
| CCRm | 0.82±0.02 | 0.89±0.01 | 0.33±0.03 | 0±0 | 0.66±0.06 | 0.56±0.04 | 0.9±0.01 | 0.8±0.03 | 0.9±0.01 | 0.73±0.05 | 0.91±0.01 | 0.81±0.03 | 0.92±0.01 |
| CCRs | 0.69±0 | 0.74±0.03 | 0±0 | 0±0 | 0.65±0.03 | 0.6±0.06 | 0.89±0.01 | 0.66±0.04 | 0.84±0.02 | 0.58±0.05 | 0.79±0.03 | 0.74±0.03 | 0.81±0 |
| CPTA Farside s | 0.64±0.11 | 0±0 | 0±0 | 0±0 | 0.58±0.06 | 0±0 | 0.86±0.02 | 0±0 | 0±0 | 0±0 | 0.86±0.02 | 0±0 | 0±0 |
| CPTA Nearside o | 0.24±0.02 | 0±0 | 0±0 | 0±0 | 0.38±0.11 | 0±0 | 0.73±0.02 | 0.15±0.01 | 0±0 | 0.35±0.08 | 0.35±0.08 | 0±0 | 0±0 |
| CPLA | 0.68±0.07 | 0.77±0.05 | 0±0 | 0.52±0.07 | 0.78±0.05 | 0.24±0.07 | 0.88±0.01 | 0.68±0.01 | 0.88±0.01 | 0.45±0.04 | 0.88±0.01 | 0.65±0.06 | 0.76±0.04 |
| CPNAO | 0.23±0.14 | 0.34±0.06 | 0±0 | 0.32±0.07 | 0.45±0.06 | 0.37±0.06 | 0.63±0.03 | 0.35±0.06 | 0.47±0.05 | 0.42±0.03 | 0.56±0.04 | 0.63±0.04 | 0.38±0.05 |
| CPTA Nearside s | 0.73±0.02 | 0±0 | 0±0 | 0±0 | 0±0 | 0±0 | 0.73±0.02 | 0±0 | 0±0 | 0±0 | 0.35±0.08 | 0±0 | 0±0 |
| CPLAs | NA | NA | NA | NA | NA | NA | NA | NA | NA | NA | NA | NA | NA |
| CPTC Farside s | 0.86±0.02 | 0±0 | 0±0 | 0±0 | 0.58±0.06 | 0±0 | 0.86±0.02 | 0±0 | 0.41±0.16 | 0.6±0.07 | 0.46±0.16 | 0±0 | 0±0 |
| CPNCO | 0.51±0.02 | 0.4±0.03 | 0±0 | 0±0 | 0.46±0.05 | 0.42±0.03 | 0.56±0.04 | 0.33±0.06 | 0.47±0.05 | 0.42±0.03 | 0.46±0.05 | 0.36±0.06 | 0±0 |
| CPNCO group | 0.63±0.04 | 0.63±0.04 | 0±0 | 0.31±0.07 | 0.6±0.02 | 0.45±0.06 | 0.67±0.02 | 0.42±0.03 | 0.59±0.03 | 0.62±0.04 | 0.67±0.02 | 0.58±0.01 | 0±0 |
| CPTC Nearside s | 0.43±0.15 | 0±0 | 0±0 | 0±0 | 0±0 | 0±0 | 0.73±0.02 | 0±0 | 0±0 | 0±0 | 0.35±0.08 | 0±0 | 0±0 |
| CCFtap | 0.22±0.03 | 0±0 | 0±0 | 0±0 | 0.8±0.01 | 0±0 | 0.79±0.01 | 0.29±0.02 | 0.38±0.01 | 0.65±0.02 | 0.39±0.02 | 0±0 | 0.57±0 |
| CCscp left | 0.13±0.02 | 0±0 | 0±0 | 0±0 | 0.3±0.02 | 0±0 | 0.35±0.02 | 0±0 | 0.41±0.01 | 0±0 | 0.13±0.01 | 0±0 | 0±0 |
| CCscp right | 0.11±0.02 | 0±0 | 0±0 | 0±0 | 0±0 | 0±0 | 0.76±0.02 | 0.14±0.02 | 0.37±0.02 | 0±0 | 0.15±0.02 | 0±0 | 0±0 |
| Pallets | 0±0 | 0±0 | 0±0 | 0±0 | 0±0 | 0±0 | 0.9±0.02 | 0±0 | 0±0 | 0±0 | 0±0 | 0±0 | 0±0 |
| Tire | 0±0 | 0±0 | 0±0 | 0±0 | 0±0 | 0±0 | 0.17±0.07 | 0±0 | 0±0 | 0±0 | 0±0 | 0±0 | 0±0 |

Table 16: Comparative assessment of the MPS performance mean ± standard deviation of each vehicle of the test pool for the full set of scenarios of the test protocol, using US statistical weights, in day light conditions.

| | MIT_POW_NIGHT_EU | | | | | | | | | | | | |
|---|---|---|---|---|---|---|---|---|---|---|---|---|---|
| MODEL | ID 1A | ID 1B | ID 2 | ID 3 | ID 4 | ID 5 | ID 6 | ID 7A | ID 7B | ID 8 | ID 9 | ID 10 | ID 11 |
| CBTA Farside | NA | NA | NA | NA | NA | NA | NA | NA | NA | NA | NA | NA | NA |
| CBNA | NA | NA | NA | NA | NA | NA | NA | NA | NA | NA | NA | NA | NA |
| CBTA Nearside | NA | NA | NA | NA | NA | NA | NA | NA | NA | NA | NA | NA | NA |
| CBNAO | NA | NA | NA | NA | NA | NA | NA | NA | NA | NA | NA | NA | NA |
| CCRm | 0.69±0.16 | 0.68±0.18 | 0±0 | 0±0 | 0.48±0.17 | 0.35±0.13 | 0.73±0.16 | 0.62±0.18 | 0.49±0.17 | 0.47±0.17 | 0.72±0.17 | 0.47±0.17 | 0.69±0.18 |
| CCRs | 0.6±0.09 | 0±0 | 0±0 | 0±0 | 0.38±0.09 | 0±0 | 0.91±0.02 | 0.36±0.09 | 0.71±0.07 | 0.42±0.10 | 0.73±0.05 | 0.43±0.10 | 0.52±0.10 |
| CPTA Farside s | 0.36±0.03 | 0±0 | 0±0 | 0±0 | 0±0 | 0±0 | 0.87±0.03 | 0±0 | 0±0 | 0±0 | 0.87±0.03 | 0±0 | 0±0 |
| CPTA Nearside o | NA | NA | NA | NA | NA | NA | NA | NA | NA | NA | NA | NA | NA |
| CPLA | 0.5±0.01 | 0.65±0 | 0±0 | 0.51±0.01 | 0.77±0.06 | 0±0 | 0.88±0.01 | 0.76±0.06 | 0.34±0.02 | 0.46±0.02 | 0±0 | 0.5±0.01 | 0.52±0.01 |
| CPNAO | 0±0 | 0.38±0.03 | 0±0 | 0±0 | 0.41±0.05 | 0±0 | 0.77±0.18 | 0±0 | 0±0 | 0±0 | 0.39±0.04 | 0.68±0.13 | 0±0 |
| CPTA Nearside s | 0.74±0.04 | 0±0 | 0±0 | 0±0 | 0±0 | 0±0 | 0.74±0.04 | 0±0 | 0±0 | 0±0 | 0.24±0.04 | 0±0 | 0±0 |
| CPLAs | 0.69±0.08 | 0.48±0.04 | 0±0 | 0.55±0.05 | 0.79±0.09 | 0±0 | 0.89±0.08 | 0.49±0.04 | 0±0 | 0.47±0.03 | 0±0 | 0.53±0.05 | 0.39±0.01 |
| CPTC Farside s | NA | NA | NA | NA | NA | NA | NA | NA | NA | NA | NA | NA | NA |
| CPNCO | 0.56±0.1 | 0.39±0.04 | 0±0 | 0±0 | 0.44±0.06 | 0±0 | 0.82±0.16 | 0±0 | 0±0 | 0±0 | 0.41±0.05 | 0±0 | 0±0 |
| CPNCO group | NA | NA | NA | NA | NA | NA | NA | NA | NA | NA | NA | NA | NA |
| CPTC Nearside s | NA | NA | NA | NA | NA | NA | NA | NA | NA | NA | NA | NA | NA |
| CCFtap | 0.26±0.04 | 0±0 | 0±0 | 0±0 | 0.5±0.11 | 0±0 | 0.86±0.05 | 0±0 | 0.27±0.04 | 0±0 | 0.47±0.09 | 0±0 | 0±0 |
| CCscp left | 0±0 | 0±0 | 0±0 | 0±0 | 0.62±0.16 | 0±0 | 0.65±0.16 | 0±0 | 0.61±0.13 | 0±0 | 0±0 | 0±0 | 0±0 |
| CCscp right | 0±0 | 0±0 | 0±0 | 0±0 | 0±0 | 0±0 | 0±0 | 0±0 | 0.56±0.11 | 0±0 | 0±0 | 0±0 | 0±0 |
| Pallets | 0±0 | 0±0 | 0±0 | 0±0 | 0±0 | 0±0 | 0.9±0.02 | 0±0 | 0±0 | 0±0 | 0±0 | 0±0 | 0±0 |
| Tire | 0±0 | 0±0 | 0±0 | 0±0 | 0±0 | 0±0 | 0.79±0.07 | 0±0 | 0±0 | 0±0 | 0±0 | 0±0 | 0±0 |

Table 17: Comparative assessment of the MPS performance mean ± standard deviation of each vehicle of the test pool for the full set of scenarios of the test protocol, using EU statistical weights, in night conditions.

| | MIT_POW_NIGHT_US | | | | | | | | | | | | |
|---|---|---|---|---|---|---|---|---|---|---|---|---|---|
| MODEL | ID 1A | ID 1B | ID 2 | ID 3 | ID 4 | ID 5 | ID 6 | ID 7A | ID 7B | ID 8 | ID 9 | ID 10 | ID 11 |
| CBTA Farside | NA | NA | NA | NA | NA | NA | NA | NA | NA | NA | NA | NA | NA |
| CBNA | NA | NA | NA | NA | NA | NA | NA | NA | NA | NA | NA | NA | NA |
| CBTA Nearside | NA | NA | NA | NA | NA | NA | NA | NA | NA | NA | NA | NA | NA |
| CBNAO | NA | NA | NA | NA | NA | NA | NA | NA | NA | NA | NA | NA | NA |
| CCRm | 0.86±0.02 | 0.86±0 | 0±0 | 0±0 | 0.65±0 | 0.48±0 | 0.89±0.01 | 0.8±0 | 0.66±0 | 0.64±0 | 0.89±0 | 0.64±0 | 0.86±0 |
| CCRs | 0.69±0 | 0±0 | 0±0 | 0±0 | 0.48±0 | 0±0 | 0.89±0.01 | 0.45±0 | 0.79±0 | 0.52±0 | 0.8±0 | 0.53±0 | 0.61±0 |
| CPTA Farside s | 0.39±0 | 0±0 | 0±0 | 0±0 | 0±0 | 0±0 | 0.86±0.02 | 0±0 | 0±0 | 0±0 | 0.86±0.02 | 0±0 | 0±0 |
| CPTA Nearside o | NA | NA | NA | NA | NA | NA | NA | NA | NA | NA | NA | NA | NA |
| CPLA | 0.52±0 | 0.65±0 | 0±0 | 0.52±0 | 0.76±0.05 | 0±0 | 0.88±0.01 | 0.75±0.05 | 0.36±0 | 0.48±0 | 0±0 | 0.52±0 | 0.53±0 |
| CPNAO | 0±0 | 0.34±0 | 0±0 | 0±0 | 0.37±0 | 0±0 | 0.63±0.04 | 0±0 | 0±0 | 0±0 | 0.35±0 | 0.55±0 | 0±0 |
| CPTA Nearside s | 0.73±0.02 | 0±0 | 0±0 | 0±0 | 0±0 | 0±0 | 0.73±0.02 | 0±0 | 0±0 | 0±0 | 0.35±0.08 | 0±0 | 0±0 |
| CPLAs | 0.6±0 | 0.45±0 | 0±0 | 0.49±0 | 0.7±0 | 0±0 | 0.82±0.01 | 0.46±0 | 0±0 | 0.44±0 | 0±0 | 0.48±0 | 0.38±0 |
| CPTC Farside s | NA | NA | NA | NA | NA | NA | NA | NA | NA | NA | NA | NA | NA |
| CPNCO | 0.46±0 | 0.35±0 | 0±0 | 0±0 | 0.39±0 | 0±0 | 0.67±0.02 | 0±0 | 0±0 | 0±0 | 0.37±0 | 0±0 | 0±0 |
| CPNCO group | NA | NA | NA | NA | NA | NA | NA | NA | NA | NA | NA | NA | NA |
| CPTC Nearside s | NA | NA | NA | NA | NA | NA | NA | NA | NA | NA | NA | NA | NA |
| CCFtap | 0.22±0 | 0±0 | 0±0 | 0±0 | 0.43±0.03 | 0±0 | 0.83±0.01 | 0±0 | 0.22±0 | 0±0 | 0.4±0.01 | 0±0 | 0±0 |
| CCscp left | 0±0 | 0±0 | 0±0 | 0±0 | 0.52±0.06 | 0±0 | 0.54±0.05 | 0±0 | 0.51±0.03 | 0±0 | 0±0 | 0±0 | 0±0 |
| CCscp right | 0±0 | 0±0 | 0±0 | 0±0 | 0±0 | 0±0 | 0±0 | 0±0 | 0.48±0.02 | 0±0 | 0±0 | 0±0 | 0±0 |
| Pallets | 0±0 | 0±0 | 0±0 | 0±0 | 0±0 | 0±0 | 0.9±0.02 | 0±0 | 0±0 | 0±0 | 0±0 | 0±0 | 0±0 |
| Tire | 0±0 | 0±0 | 0±0 | 0±0 | 0±0 | 0±0 | 0.79±0.07 | 0±0 | 0±0 | 0±0 | 0±0 | 0±0 | 0±0 |

Table 18: Comparative assessment of the MPS performance mean ± standard deviation of each vehicle of the test pool for the full set of scenarios of the test protocol, using US statistical weights, in night conditions.

## Appendix C

**Number of tests per vehicle**

Across the thirteen vehicles in the test pool, the average number of tests executed and analyzed in this manuscript per vehicle is 81 tests, representing approximately 26% of the full test protocol (see Table 2). The upper and lower bound of tests executed are 161 tests for the best performer (72% of the full test matrix) and 41 tests for the worst performer (18% of full the test matrix).

| Vehicle | Test protocol (expected) | Test protocol (executed) | Judged[9] | % |
|---|---|---|---|---|
| ID 1A | 224 | 105 | 5 | 49% |
| ID 1B | 224 | 57 | 9 | 29% |
| ID 2 | 224 | 26 | 15 | 18% |
| ID 3 | 224 | 35 | 12 | 21% |
| ID 4 | 224 | 83 | 5 | 39% |
| ID 5 | 224 | 29 | 12 | 18% |
| ID 6 | 224 | 161 |  | 72% |
| ID 7A | 224 | 61 | 9 | 31% |
| ID 7B | 224 | 100 | 8 | 48% |
| ID 8 | 224 | 56 | 8 | 29% |
| ID 9 | 224 | 91 | 4 | 42% |
| ID 10 | 224 | 62 | 6 | 30% |
| ID 11 | 224 | 65 | 10 | 33% |

Table 2C: *Expected* vs. *executed* tests for each single vehicle, including *judged* test, and relative percentage of completion rate.

## CRediT authorship contribution statement

**Luigi Di Lillo**: Writing – review & editing, Writing – original draft, Visualization, Validation, Supervision, Project administration, Methodology, Formal analysis, Data curation, Conceptualization.
**Andrea Triscari**: Writing – review & editing, Writing – original draft, Visualization, Validation, Methodology, Formal analysis, Data curation, Conceptualization. **Xilin Zhou**: Writing – review & editing, Writing – original draft, Visualization, Methodology, Conceptualization. **Robert Dyro**: Writing – original draft; Conceptualization; Definition and implementation of realism score; Data curation.
**Ruolin Li**: Writing – original draft; Evaluation of realism score; Data curation. **Marco Pavone**: Writing – original draft; Project administration; Definition of realism score.

---

[9] "Judged" refers to tests that were judged by the test engineer as a failure when not performed due to i) either the vehicle had failed in a lower speed test in the same scenario or, ii) in case of night tests, when the vehicle failed in a similar scenario during the day.